\documentclass[journal,onecolumn]{IEEEtran}
\IEEEoverridecommandlockouts
\usepackage{cite}
\usepackage{amsmath,amssymb,amsfonts}
\usepackage{algorithmic}
\usepackage{graphicx}
\usepackage{textcomp}
\usepackage{xcolor}
\usepackage{tabularx}
\usepackage{booktabs}
\usepackage{multirow}
\usepackage{textcomp}
\usepackage{bbm}
\usepackage{xcolor}
\usepackage{url}
\usepackage{soul}
\usepackage{glossaries}
\usepackage{makecell}
\usepackage{hyperref}
\setacronymstyle{long-short}
\def\BibTeX{{\rm B\kern-.05em{\sc i\kern-.025em b}\kern-.08em
    T\kern-.1667em\lower.7ex\hbox{E}\kern-.125emX}}

\begin{document}

\title{Braille Letter Reading: A Benchmark for Spatio-Temporal Pattern Recognition on Neuromorphic Hardware \\
}

\author{
\IEEEauthorblockN{Simon F M{\"u}ller-Cleve\,$^{1,*}$, 
Vittorio Fra\,$^{2}$, 
Lyes Khacef\,$^{3}$, 
Alejandro Peque\~no-Zurro\,$^{4}$, \\
Daniel Klepatsch\,$^{5,6}$,
Evelina Forno\,$^{2}$, 
Diego G Ivanovich\,$^{5,6}$, 
Shavika Rastogi\,$^{7,8}$, \\
Gianvito Urgese\,$^{2}$, 
Friedemann Zenke\,$^{9,10}$, 
Chiara Bartolozzi\,$^{1}$} \\

\IEEEauthorblockA{\textit{$^{1}$Istituto Italiano di Tecnologia, Event-driven perception in robotics, Genoa, Italy,
$^{2}$Politecnico di Torino, EDA Group, Torino, Italy,
$^{3}$Bio-Inspired Circuits and Systems Lab, Zernike Institute for Advanced Materials, Groningen Cognitive Systems and Materials Center, University of Groningen, Groningen, Netherlands,
$^{4}$Laboratory of Neural Computation, Istituto Italiano di Tecnologia, Genova , Italy,
$^{5}$Silicon Austria Labs, JKU LIT SAL eSPML Lab, Austria,
$^{6}$Johannes Kepler University Linz, JKU LIT SAL eSPML Lab, Austria,
$^{7}$International Centre for Neuromorphic Systems, Western Sydney University, Australia,
$^{8}$ Biocomputation Research Group, University of Hertfordshire, UK,
$^{9}$Friedrich Miescher Institute for Biomedical Research, Neurobiology, Basel, Switzerland,
$^{10}$University of Basel, Basel, Switzerland}}}


\newacronym{ann}{ANN}{Artificial Neural Network}
\newacronym{api}{API}{Application Programming Interface}
\newacronym{bptt}{BPTT}{Backpropagation Through Time}
\newacronym{cdc}{CDC}{Capacitance to Digital Converter}
\newacronym{cnn}{CNN}{Convolutional Neural Network}
\newacronym{cuba}{CUBA}{current-based}
\newacronym{tdm}{TDM}{Time-division Multiplexing}
\newacronym{dvs}{DVS}{Dynamic Vision Sensor}
\newacronym{fc}{FC}{Fully-Connected}
\newacronym{fcn}{FCN}{Fully Convolutional Network}
\newacronym{ffsnn}{FFSNN}{Feedforward Spiking Neural Network}
\newacronym{gap}{GAP}{Global Average Pooling}
\newacronym{hpo}{HPO}{Hyperparameters Optimization}
\newacronym{isi}{ISI}{interspike interval}
\newacronym{isi}{ISI}{Interspike Interval}
\newacronym{lif}{LIF}{Leaky Integrate and Firing Neuron Model}
\newacronym{lstm}{LSTM}{Long Short-Term Memory}
\newacronym{mse}{MSE}{Mean Squared Error}
\newacronym{mse}{MSE}{Mean Squared Error}
\newacronym{mtb}{MTB}{Microcontroller Board}
\newacronym{mts}{MTS}{Multivariante Time Series}
\newacronym{nni}{NNI}{Neural Network Intelligence}
\newacronym{os}{OS}{Operating System}
\newacronym{pca}{PCA}{Principal Component Analysis} 
\newacronym{pcb}{PCB}{Printed Circuit Board}
\newacronym{prelu}{PReLU}{Parametric Rectified Linear Unit}
\newacronym{resnet}{ResNet}{Residual Neural Network}
\newacronym{rsnn}{RSNN}{Recurrent Spiking Neural Network}
\newacronym{simd}{SIMD}{Single-Instruction Multiple-Data}
\newacronym{snn}{SNN}{Spiking Neural Network}
\newacronym{soc}{SoC}{System-on-Chip}
\newacronym{som}{SOM}{System-on-Module}
\newacronym{svm}{SVM}{Support-Vector Machine}
\newacronym{tcnn}{TCNN}{Time-CNN}
\newacronym{ttc}{TTC}{Time-to-Classify}

\maketitle


\begin{abstract}
Spatio-temporal pattern recognition is a fundamental ability of the brain which is required for numerous real-world activities. Recent deep learning approaches have reached outstanding accuracies in such tasks, but their implementation on conventional embedded solutions is still very computationally and energy expensive. Tactile sensing in robotic applications is a representative example where real-time processing and energy efficiency are required. Following a brain-inspired computing approach, we propose a new benchmark for spatio-temporal tactile pattern recognition at the edge through Braille letter reading. We recorded a new Braille letters dataset based on the capacitive tactile sensors of the iCub robot's fingertip. We then investigated the importance of spatial and temporal information as well as the impact of event-based encoding on spike-based computation. Afterward, we trained and compared feedforward and recurrent Spiking Neural Networks (SNNs) offline using Backpropagation Through Time (BPTT) with surrogate gradients, then we deployed them on the Intel Loihi neuromorphic chip for fast and efficient inference. We compared our approach to standard classifiers, in particular to the Long Short-Term Memory (LSTM) deployed on the embedded NVIDIA Jetson GPU, in terms of classification accuracy, power, and energy consumption together with computational delay. Our results show that the LSTM reaches $\sim$97\% of accuracy, outperforming the recurrent SNN by $\sim$17~\% when using continuous frame-based data instead of event-based inputs. However, the recurrent SNN on Loihi with event-based inputs is $\sim$500 times more energy-efficient than the LSTM on Jetson, requiring a total power of only $\sim$30~mW. This work proposes a new benchmark for tactile sensing and highlights the challenges and opportunities of event-based encoding, neuromorphic hardware, and spike-based computing for spatio-temporal pattern recognition at the edge. \\

\textbf{\textit{Keywords:}} 
\textit{Spatio-temporal pattern recognition, Braille reading, tactile sensing, event-based encoding, neuromorphic hardware, spiking neural networks, benchmarking.}

\end{abstract}


\section{Introduction} \label{Introduction}
Touch, or tactile perception, is a critical component of sensorimotor activity~\cite{romo2001touch}.
Uniquely among the senses, in many situations deliberate action by the subject is required to experience tactile feedback: this is known as \textit{active touch}~\cite{prescott2011activetouch}. Active use of touch is important for blind or visually impaired subjects, as visual perception may need to be aided or replaced by tactile and auditory information~\cite{Bach-y-Rita04}. One example of active touch is the reading of Braille letters, where fingers slide over lines of characters that are typeset into surfaces as a set of 1-6 embossed dots arranged in a 2$\times$3 matrix. In Braille reading, characters are read sequentially, unlike print reading, where entire words or groups of words can be perceived simultaneously by the eye. However, expert users can significantly speed up their Braille reading as they learn to identify various lexical, perceptual, and contextual clues~\cite{martiniello2020association}, achieving optimal reading speeds of at most 80-120 words per minute~\cite{bola2016Braille}, which is about half the average silent reading rate for adults in English~\cite{brysbaert2019many}.

The sequential and time-dependent nature of Braille reading makes it an excellent benchmark for machine learning applications involving time-varying signals since Braille text can be represented by sequential data acquired by moving tactile sensors. Optical classification applications for Braille have been developed both in machine learning~\cite{kawabe2019experimental, li2010optical} and deep learning~\cite{hsu2020Braille, shokat2020deep}. However, the image-based approach for Braille recognition requires good quality Braille samples and accurate preprocessing. Braille documents are characterized by the lack of any contrast in color between the text and the background and may require specific lighting and camera settings to produce accurate results~\cite{li2014deep}. On the other hand, automatic Braille reading relying on tactile sensors requires adding tactile sensors to robots that can slide their sensors over surfaces. The high number of sensors necessary to obtain acceptable performance adds a high overhead in terms of area, power consumption, and communication latency.

A possible solution to these problems is the inherent data encoding capabilities and sparse transmission of neuromorphic event-driven sensing~\cite{bartolozzi2016robots}. Event-driven sensors only transmit signals when a change has been detected in their sensory space, reducing communication and processing costs~\cite{bartolozzi2017event}. The event-driven domain~\cite{lichtsteiner2008128} features binary, time-discrete events, avoiding the continuous polling of sensor readouts. This is especially desirable in the domain of touch, because tactile perception is naturally sparse from a temporal and spatial perspective, with only local correlation, e.g. multiple local neighbors of receptors, or sensors, are activated by the same event. Tactile events are perceived for a limited time, as long as the stimulus is applied, and in a localized part, or patch, of the sensor. When no stimulus is present, the tactile system can be considered at rest.

While other event-driven neuromorphic sensors such as the \gls{dvs}~\cite{conradt2009embedded} and the silicon cochlea~\cite{chan2007aer} have attracted much interest from researchers, leading to specialized data pipelines and standardized benchmarks such as the \href{https://research.ibm.com/interactive/dvsgesture/}{DVS gesture recognition dataset} and TIDigits~\cite{leonard1993tidigits} dataset, there have been comparatively few such developments in the field of touch. Among the scarce number of examples of tactile classifiers exploiting tactile neuromorphic sensors~\cite{friedl2016human, rongala2015neuromorphic}, See et al. proposed the Spiking Tactile MNIST (ST-MNIST) dataset of handwritten digits~\cite{see2020stmnist} obtained by writing on a neuromorphic tactile sensor array. Consequently, the information in the spike patterns is mostly spatial, and a feedforward \gls{cnn} reaches the best accuracy when summing up all the spikes to get a "tactile image". Bologna et al. proposed a neuroengineering framework for robotic applications, including spatio-temporal event coding, probabilistic decoding, and closed-loop motion policy adaptation for active touch. The system was benchmarked on Braille, proving effective modulation of fingertip kinematics depending on character complexity. Classification of the signals recorded by a fingertip sensor mounted on a robotic arm was performed using a subset of 7 Braille characters, achieving a recognition rate of (89~$\pm$~5.3)~\%~\cite{bologna2013closed,pinoteau2012closed}.

In this paper, we propose a development path applicable to neuromorphic tasks in the tactile domain. The proposed method, while developed and tested on tactile output obtained from capacitive sensors, can be generalized to a whole class of time-dependent data such as audio streams, inertial sensor outputs, and temperature or voltage monitoring, to name just a few. Given the inherently time-dependent nature of its information content, we selected the Braille reading problem as a benchmark, for which we designed an end-to-end event-based neuromorphic classification procedure. We acquired a reliable dataset, applied a widely adopted and well know encoding technique such as the one provided by a sigma-delta modulator, and finally performed classification relying on a neuro-inspired approach, using \glspl{ffsnn} and \glspl{rsnn} models, both of them implemented in software and hardware. For the hardware implementation, we selected the NVIDIA Jetson, a modular embedded GPU platform, and the Intel Loihi, a dedicated neuromorphic chip, and benchmarked them against standard classifiers in terms of classification accuracy, average power usage, energy consumption, and computation delay during inference.

Using a comprehensive analysis accounting for multiple aspects such as the information loss introduced by the encoding technique, the accuracy of the classifiers, and the power consumption on different hardware with different models, we demonstrate that Braille reading can be performed in a highly energy-efficient way by using event-based data and deploying \glspl{snn} on dedicated neuromorphic hardware.


\section{Methods} \label{Methods}
\begin{figure}
    \centering
    \includegraphics[width=0.95\textwidth]{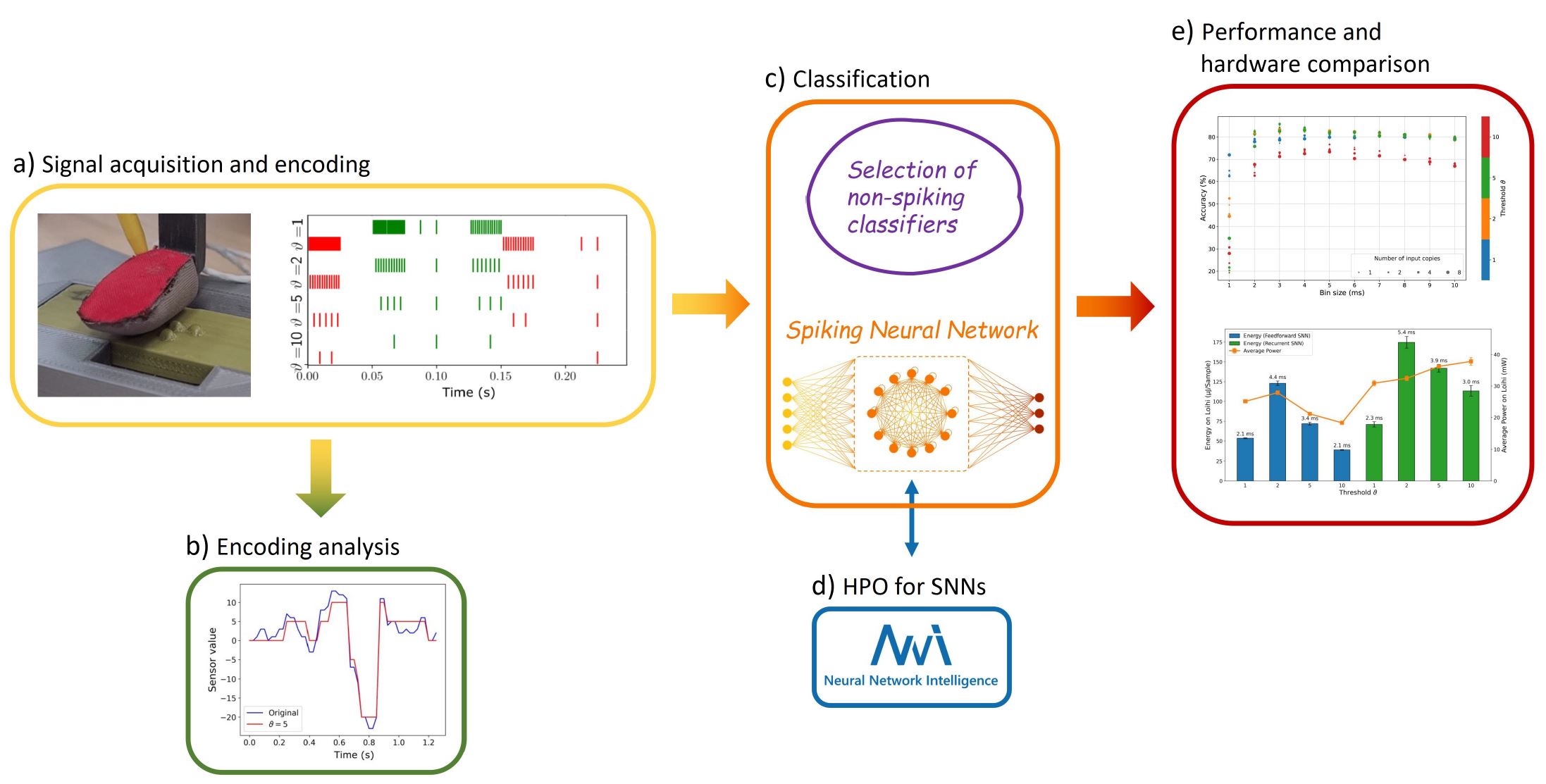}
    \caption{The workflow is summarized in five steps. Dataset acquisition and signal encoding (a) with analysis of information content and reconstruction loss (b). Different non-spiking classifiers are identified and employed to produce references for the proposed \gls{rsnn} (c), with the latter undergoing a hyperparameter optimization (d). Finally, performances are evaluated, accounting for different metrics and hardware implementations.}
    \label{fig:overall_pipeline}
\end{figure}

\subsection{The dataset} \label{methods:the_dataset}

The \href{https://www.forcedimension.com/products}{Omega.3} robot was used to slide a sensorized fingertip~\cite{jamali2015icubfingertip} over 3D printed Braille letters from \textit{`A'} to \textit{`Z'} as well as \textit{`space'} at a controlled speed and position~\footnote{The robot is controlled and the sensor response is stored using YARP on a DELL XPS 15 laptop running Ubuntu 20.04 LTS.}.

\begin{figure}
    \centering
    \includegraphics[width=0.9\textwidth]{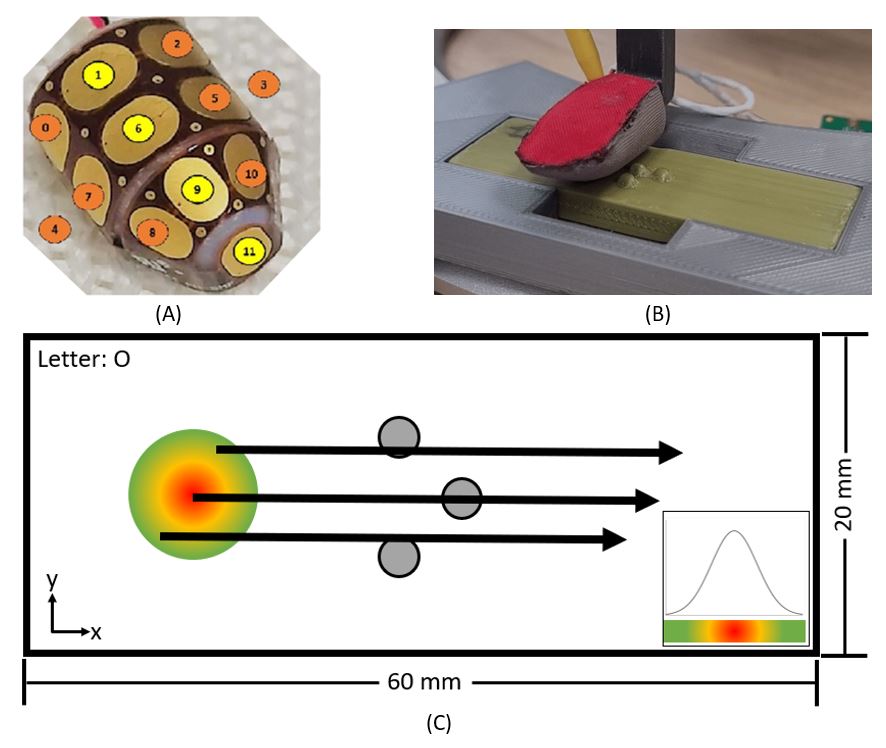}
    \caption{The inner part of the fingertip is composed of 12 capacitance plates shown in (a), wrapped by a three-layer fabric and slid over the Braille letters with a constant sliding distance and velocity (b). The start position was varied following a Gaussian distribution (c).}
    \label{fig:fingertip_and_setup}
\end{figure}

The size of the Braille letter was chosen to match the spatial distribution of the fingertip so that the full letter can be detected by a single sliding movement. The sliding distance (15.5 mm), the sliding velocity (20 mm/s), and the distance to the flat surface of the plate were held constant. The start position varied following a Gaussian distribution to include spatial-temporal variability between each repetition, illustrated in Fig.~\ref{fig:fingertip_and_setup}c. Each letter was recorded 200 times with a sampling frequency of 40~Hz. The capacitance value is encoded in 8-bit, leading to a range from 255-0 with a minimal change of 1, with 255 being in rest and 0 for maximum load. For convenience, the encoding was inverted in software being 0 the resting state, and 255 the maximum load of the sensor capacity.

\subsubsection{Event-based encoding} \label{methods:encoding}
\begin{figure}
    \centering
    \includegraphics[width=1\textwidth]{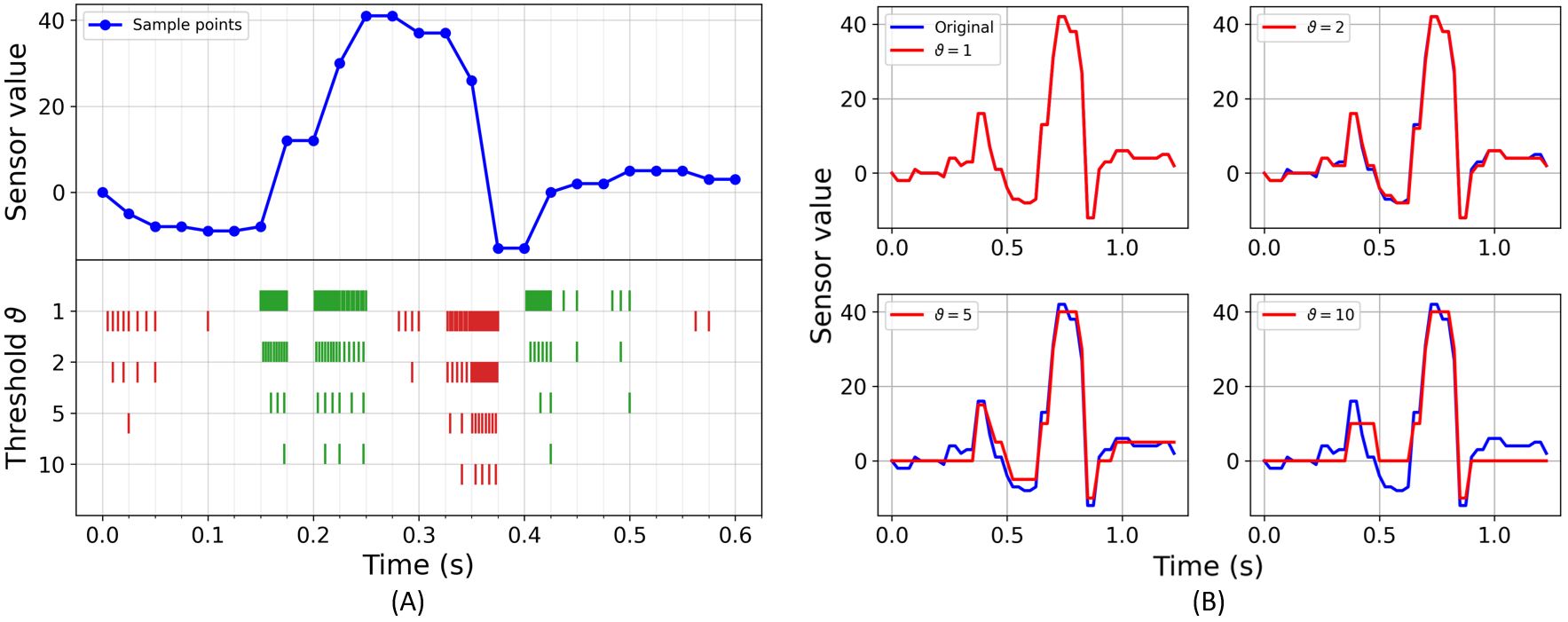}
    \caption{Event-based encoding and reconstruction of a sample: (a) Sensor reading sequence of a sample letter along with spikes generated using sigma-delta modulation. The upper part represents 600~ms of a sequence of sensor readings of a single taxel during sliding. The bottom part shows the generated events for the ON (green color) and OFF (red color) channels for increasing threshold values, leading to decreasing numbers of events. (b) Reconstructed sequence from event-based data compared with the original sequence for a full letter sequence. Each plot represents the reconstruction with a different threshold of the same frame-based sequence. Increasing thresholds increase the compression, but also increase the reconstruction error.}
    \label{fig:encode_reconstruct}
\end{figure}

The objective of this work is to explore the potential of an end-to-end neuromorphic system for tactile perception with event-based communication (sensor level), asynchronous processing (hardware level) and spike-based computing (algorithmic level). However, there is to the best of our knowledge no available event-based tactile sensor today. Therefore, we emulate the output of such a sensor by encoding the frame-based data into temporally sparse streams of events (i.e. spikes) using a sigma-delta modulator ($\Sigma\Delta$ modulator)~\cite{rosa_2011}. At threshold ($\vartheta$) crossings, ON or OFF digital events are generated for increase or decrease of pressure, respectively~\cite{bartolozzi2017event}, as indicated in Fig.~\ref{fig:encode_reconstruct}a. Each original stream of frames is converted in an offline preprocessing step from a 12-taxel time sequence to 24 binary event-based channels, emulating an event-based tactile sensor.

The maximum re-sampling frequency is given by the precision of the stored 64-bit float variable. Given the sensor encoding regime and the sampling frequency, the precision of the conversion is enough to encode the data with a minimum change in the sensor value of $\vartheta=1.02\mathrm{E}-11$ without losing information from the frame-based data. A threshold value $\vartheta=1$ corresponds to the highest implemented precision and no information loss. Increasing values correspond to increasing sparsity, lower data rate, and improved efficiency at the cost of lower accuracy and information loss, as shown in Fig.~\ref{fig:encode_reconstruct}b.

\subsection{Standard classifiers}\label{methods:standard_classifiers}
Non-event-driven approaches such as linear classifiers (e.g. \gls{svm}), time-series classifiers, and \gls{lstm} were used on frame-based signals available in the dataset as baselines for more traditional strategies independent of neuromorphic, event-based approach. Additionally, \gls{lstm} was used for event-based data as a benchmark for \glspl{snn}, too.

\subsubsection{Time-series classifiers for frame-based data} \label{method:time_series_classifiers}
We used standard time-series classifiers proven to work with time-variant datasets: \gls{fcn}~\cite{wang2017ijcnn}, \gls{resnet}~\cite{wang2017ijcnn, geng2018cscnn}, Encoder~\cite{santiago2018encoder}, \gls{tcnn}~\cite{zhao2017cnntimeseries}, and Inception~\cite{hassen2020inceptiontime}, available as \href{https://github.com/hfawaz/dl-4-tsc}{implementation on GitHub}~\cite{ismail2019deeplearning, hassen2020inceptiontime}. Please find more detailed information in the Supplementary Material~1.1

While these networks were specifically designed for time series, they are still based on standard feedforward structures (e.g. fully-connected layers, convolutional layers). Their lack of internal memory and recurrence requires that a whole time series is presented to the network all at once, instead of only the data from one timestep. Consequently, during inference, a buffer with the length of a time-series needs to be filled to perform classification. In case of multiple overlapping time windows, even more, buffers are required. This leads to memory overhead and classification delay compared to networks with a recurrent structure.

\subsubsection{Long short-term memory} \label{methods:lstm}

Given the time-dependent characteristic of the dataset, we implemented a recurrent neural network. This type of network, in contrast to feedforward networks, incorporates an internal loop that allows temporal information to persist, and therefore is the natural choice for sequential data, but suffers from vanishing gradients for sequences characterized by long-term dependencies. Such problem is avoided by using \gls{lstm} architectures~\cite{hochreiter1997lstm}, where the cell state $C_{t}$ holds long-term information. Additionally, \glspl{lstm} can add or remove information from the cell state by \textit{gates}.\\
The architecture chosen for the Braille dataset consists of a single layer \gls{lstm} with 228 hidden nodes, followed by a regular fully-connected layer of 228$\times$27 output neurons that performs the classification, giving a total number of 225,975 trainable parameters. The choice was made to have a number of trainable parameters as close as possible to that of the best performing \gls{rsnn}, obtained through a two-step \gls{hpo} procedure as described in the following, for a fair comparison. The method for calculating the number of trainable parameters is reported in Supplementary Material~1.2,~Equ.~S.1.

\subsection{Spiking neural networks for event-based data} \label{methods:snn} 

We designed a two-layer \gls{rsnn} adopted from~\cite{cramer2020heidelberg} and~\cite{zenke2021remarkable} to perform classification on the dataset encoded as an event-stream with four different thresholds, to achieve a quantitative comparison of the different possible strategies suitable to effectively deal with time-based Braille reading signals. We used the \gls{cuba} \gls{lif} neuron model written in continuous form as

\begin{equation}
\label{eq:snn1}
    \tau_{mem} \frac{\mathrm{d}U_i^{(l)}}{\mathrm{d}t}=-(U_i^{(l)} - U_{rest}) + RI_i^{(l)}
\end{equation}

with \(U_i\) being the membrane potential of neuron \(i\) (hidden state) in layer \(l\), \(U_{rest}\) being the resting potential, $\tau_{mem}$ the membrane time constant, \(R\) the input resistance, and \(I_i\) being the input current defined as

\begin{equation}
\label{eq:snn2}
    \frac{\mathrm{d}I_i}{\mathrm{d}t}=\frac{I_i(t)}{\tau_{syn}} + \sum_jW_{ij}S_j^{(0)}(t) + \sum_jV_{ij}S_j^{(1)}(t)
\end{equation}

with $\tau_{syn}$ being the synapse decay time constants, $S_j(l)$ the spike train of the $j$th neuron at the $l$th layer, $W_{ij}$ being the forward, and $V_{ij}$ the recurrent connection's weights.

\begin{equation}
\label{eq:snn3_a}
    I_i^{(l)}(t)=\alpha I_i^{(l)}(t-1) + \sum_j W_{ij} \cdot S_j(t)
\end{equation}

\begin{equation}
\label{eq:snn3_b}
    U_i^{(l)}(t)=(\beta U_i^{(l)}(t-1) + I_i^{(l)}(t)) \cdot  (1.0 - U_{reset})
\end{equation}

with $\beta=\mathrm{exp}(\frac{-time\_bin\_size}{\tau_{mem}})$ being the voltage decay constant, $\alpha=\mathrm{exp}(\frac{-time\_bin\_size}{\tau_{syn}})$ the current decay constant, and $I_i^{(l)}$ the synaptic input current from neuron $i$ in layer $l$ multiplied by the input resistance $R=1\Omega$ for convenience and $U_{reset}$ the reset after eliciting an event.

The error is propagated throughout the entire network, unrolled in time, using \gls{bptt}. To perform supervised learning the feedforward and recurrent weight matrices $W_{ij}$ and $V_{ij}$ change following a given loss $\mathcal{L}$

\begin{equation}
\label{eq:snn4}
    W_{ij} \leftarrow W_{ij} - \eta \frac{\partial \mathcal{L}}{\partial W_{ij}} \quad \textrm{and} \quad V_{ij} \leftarrow V_{ij} - \eta \frac{\partial \mathcal{L}}{\partial V_{ij}}
\end{equation}

with the learning rate \(\eta\). To use a binary step function \(\Theta(x)\) in the forward pass (inference), whose derivative is zero everywhere except the zero crossing where it becomes infinite, we use the partial derivative (the gradient) of a fast sigmoid function \(\sigma(x)\)

\begin{equation}
\label{eq:snn5}
    \sigma(U_i^{(l)})=\frac{U_i^{(l)}}{1 + \lambda |U_i^{(l)}|}
\end{equation}

in the backward pass (training), the surrogate gradient, and prevent vanishing issues. Whereas \(\Theta(x)\) is invariant to multiplicative re-scaling, \(\sigma(x)\) needs the introduction of the scale parameter $\lambda$  being part of the hyperparameter optimization.

To compute the gradients we are using the capabilities to over-loading the derivative of spiking nonlinearity with a differential function in custom PyTorch~\cite{neftci2019surrogategradient, zenke2021remarkable}.

For the loss, we apply the cross entropy to the active readout layer \(l=L\). For data with \(N_{batch}\) samples and \(N_{class}\) classes it is formalized as

\begin{equation}
\label{eq:snn6}
    \mathcal{L}=- \frac{1}{N_{batch}} \sum_{s=1}^{N_{batch}} \mathbbm{1}(i=y_2) \cdot \mathrm{log}\left\{\frac{\mathrm{exp}\bigg(\sum_{n=1}^T S_i^{(l)} [n]\bigg)}{\sum_{i=1}^{N_{class}} \mathrm{exp}\bigg(\sum_{n=1}^T S_i^{(l)} [n]\bigg)} \right\}
\end{equation}

whereas $n$ stands for the time step. At last we have to define the \(\mathcal{L}_1\) and \(\mathcal{L}_2\) regularization loss function.

\begin{equation}
\label{eq:snn7}
    \mathcal{L}_1=\frac{s_l}{N_{batch}+N} \sum_{s=1}^{N_{batch}} \sum_{i=1}^{N} \mathrm{max}\bigg\{0, \frac{1}{T} \sum_{n=1}^T S_i^{(l)} [n] - \theta_l \bigg\}
\end{equation}

representing a per neuron lower threshold spike count regularization with strength \(s_l\) and threshold \(\theta_l\), and

\begin{equation}
\label{eq:snn8}
    \mathcal{L}_2=\frac{s_u}{N_{batch}} \sum_{s=1}^{N_{batch}} \bigg[\mathrm{max}\bigg\{0, \frac{1}{N} \sum_{i=1}^N \sum_{n=1}^T S_i^{(l)} [n] - \theta_u \bigg\}\bigg]^2
\end{equation}

being an upper threshold mean population spike count regularization with strength \(s_u\) and threshold \(\theta_u\). Finally, the total loss is summarized by 
 
\begin{equation}
\label{eq:snn9}
    \mathcal{L}_\mathrm{tot}=\mathcal{L} + \mu_1\mathcal{L}_1 + \mu_2\mathcal{L}_2
\end{equation}

with $\mu$ as a scaling factor and minimized using the Adamax optimizer~\cite{adam2017autodiff}.

To implement and simulate \glspl{snn} based on this model in PyTorch, we need to account for a time binning step for the input event stream. Although aiming to work with asynchronous and sparse event-based data, fixed frame lengths had to be defined to properly simulate algorithmic time steps in the domain of clock-driven, conventional hardware like CPUs and GPUs. Time binning was performed by subdividing the time extracted from the signal recordings ($T_{rec}$) into $T$ chunks, with $T$ defined as $T=\mathrm{int}(T_{rec}/time\_bin\_size)$ and the quantity $time\_bin\_size$ introduced as an additional hyperparameter of the \gls{hpo}. Then, by iterating over the encoded signal with a stride equal to $time\_bin\_size$, a value of 1 was assigned whenever at least one spike was found, otherwise a 0. The winning neuron in the output layer is the one with the highest spike count after a trial.

\subsection{Hyperparameter optimization} \label{methods:hpo}

For each event stream produced from the original frame-based signal by applying a specific threshold value, a tailored \gls{rsnn} was obtained by adapting the parameter optimization procedure introduced in~\cite{fra2022human}. The \gls{hpo} was performed by means of the \textit{Anneal} algorithm in the \gls{nni} toolkit, using the parameters listed in Tab.~\ref{tab:hpo}, over 600 trials. To partially mitigate the impact of local minima~\cite{forno2019parallel}, two evenly spaced random reinitializations of the tuner were performed during each experiment. An 80/20 train-test split was used and all trials were composed of 300 training epochs with intermediate results, for both training and test, at the end of each epoch. Test accuracy was defined as the optimization objective of the \gls{hpo} experiments, and its highest value was extracted at the end of each trial. The choice of selecting test accuracy as a reference value to be optimized was taken to account for possible overfitting.

Following the annealing-based procedure, we performed a further exploration of a portion of the initial search space through a grid search on the two most relevant hyperparameters from the energy consumption perspective, namely $time\_bin\_size$ and $nb\_input\_copies$ since they determine the number of operations that need to be computed per inference.

As the outcome of such a two-step \gls{hpo} procedure, an optimized network for each threshold value used in the sigma-delta encoding was obtained. All of these \gls{rsnn}s were composed of a recurrent, fully connected hidden layer containing 450 \gls{lif} neurons and an output layer composed of 28 \gls{lif} neurons. The number of output neurons was defined to account for an extra class, in addition to the 27 defined by the letters, suitable to identify, given future possible online implementations, edge cases such as missing contact between the fingertip and the letters. The input layer was instead part of the optimization, with its number of input neurons defined as $2 \cdot n\_taxels \cdot nb\_input\_copies$ where $2$ covers the event polarity, $n\_taxels$ was given by the 12 sensors in the robotic fingertip, and $nb\_input\_copies$ was to be optimized. A batch size of 128 and a learning rate $\eta=0.0015$ were adopted for all the networks.

\begin{table}
    \caption{Description of the hyperparameters included in the search space for the \gls{hpo} procedure.\\}
    \label{tab:hpo}
    \centering
    \begin{scriptsize}
        \begin{tabular}{>{\centering}m{0.3\linewidth}>{\centering\arraybackslash}m{0.5\linewidth}}
	        \textbf{Hyperparameter} & \textbf{Description} \\
	        \toprule
	        \textbf{scale}, $\lambda$ & Steepness of surrogate gradient \\
	        \midrule
	        \textbf{time\_bin\_size} & Time binning of the encoded input \\
	        \midrule
	        \textbf{nb\_input\_copies} & Copies of the encoded signals provided to the input layer \\
	        \midrule
	        \textbf{tau\_mem}, $\tau_{mem}$ & Decay time constant of the membrane \\
	        \midrule
	        \textbf{tau\_ratio} & Ratio between the membrane and the synapse ($\tau_{syn}$) decay time constants \\
	        \midrule
	        \textbf{fwd\_weight\_scale} & Scaling factor for weight initialization of the forward connections ($W_{ij}$) \\
	        \midrule
	        \textbf{weight\_scale\_factor} & Scaling factor for weight initialization of the recurrent connections ($V_{ij}$) \\
	        \midrule
	        \textbf{reg\_neurons}, $\mu_1$ & Scaling factor for the regularization on the number of spikes per neuron \\
	        \midrule
	        \textbf{reg\_spikes}, $\mu_2$ & Scaling factor for the regularization on the total number of spikes \\
	        \bottomrule
        \end{tabular}
    \end{scriptsize}
\end{table}

\subsection{Hardware implementation} \label{methods:hardware}

Beyond the algorithmic evaluation, we determined key performance metrics that are relevant to real-world deployment by implementing the networks on different hardware platforms. These metrics covered power usage, energy consumption, and computational delay, which allowed us to conclude the deployment feasibility in real-world scenarios. Considering the platform-related factors of high integration and availability, and our ultimate goal of deploying the algorithms in a real-world environment on robots, we targeted the NVIDIA Jetson Xavier NX, a commercially off-the-shelf available computing platform equipped with a \gls{soc} that integrates a CPU and GPU, and the Intel Loihi, a neuromorphic processor dedicated to accelerating \glspl{snn}.

\subsubsection{NVIDIA Jetson Xavier NX}

The NVIDIA Jetson is a product family of compact and embedded computation platforms mainly targeted toward edge AI. The Xavier NX is the most powerful model among its compact 260-pin SO-DIMM modules. Despite being a general-purpose platform, it is similar to common machine learning workstations in terms of architecture and software. We used it to run all algorithms and compare the different standard time-series classifiers, as well as evaluate the differences between conventional algorithms and event-based algorithms on off-the-shelf hardware. Furthermore, the inference metrics give an outlook on what performance can be expected during deployment with the same hardware.

Execution time was measured by using the native functionality offered by the Linux \gls{os}. Power usage was measured by utilizing the module's onboard  \href{https://docs.nvidia.com/jetson/archives/r34.1/DeveloperGuide/text/SD/PlatformPowerAndPerformance/JetsonXavierNxSeriesAndJetsonAgxXavierSeries.html#software-based-power-consumption-modeling}{INA3221 power monitor}, polling the system's main power rail with a fixed interval of 50 ms. The power monitor also measures a CPU/GPU and a \gls{soc} power rail. But due to the lack of public information on what components exactly these rails supply, and also due to a productive application requiring a full system instead of just single core components, the main power rail was chosen for comparison.

The general procedure for measuring the performance metrics was performed as follows: initially, the whole dataset was loaded into memory, followed by the loading of the model and its trained weights. Next, a warmup of the system was performed by letting the model predict the whole dataset in batches of 64 for six times. Its purpose was to fill the caches and to avoid additional library load times during the actual inference. Afterward, the recordings of the execution time and power values were started, which was immediately followed by the inference. Like in the warmup, the whole dataset was predicted six times, but with the difference that a batch size of 1 was used to simulate how the system would behave in a real-world scenario where single samples are predicted consecutively. For our \glspl{snn} the number of samples during warmup and inference was reduced to 750 each due to timing constraints. Finally, after the inference was done, the recordings were stopped and the following metrics were evaluated:

\begin{itemize}
    \item Inference time (i.e. computational delay) per sample, which is the total inference time divided by the number of samples processed.
    \item Minimum, Maximum, and Average power usage over total inference time as well as per sample.
    \item Total and per sample energy consumption. The total energy is calculated by multiplying and accumulating each power measurement with a polling interval of 50 ms. Energy per sample is given by dividing the total energy by the number of samples processed.
\end{itemize}

Furthermore, to get more significant results, the above procedure was repeated three times per network to eliminate possible outliers and also performed for each available power mode on the Jetson, whereas the results reported are from power mode 4.

\subsubsection{Intel Loihi}
Intel's Loihi \cite{davies2018loihi} is a fully digital neuromorphic research processor. Each Loihi chip hosts 128 neuron cores, where every neural core can run up to 1,024 \gls{cuba} \gls{lif} neurons by \gls{tdm}. The Loihi neuron's equations for the current and voltage compartments are

\begin{equation}
\label{eq:loihi_current}
    I_i(t)=I_i(t-1) \cdot (2^{12} - \delta_i^I) \cdot 2^{-12} + 2^6 \cdot \sum_j w_{ij} \cdot s_j(t) 
\end{equation}

\begin{equation}
\label{eq:loihi_voltage}
    U_i(t)=U_i(t-1) \cdot (2^{12} - \delta_i^U) \cdot 2^{-12} + I_i(t)
\end{equation}

where $t$ is the algorithmic time step, $I_i(t)$ and $U_i(t)$ are the current and voltage of neuron $i$, $\delta_i^I$ and $\delta_i^U$ are the current and voltage decay constants, $w_{ij}$ is the synaptic weight from neuron $j$ to $i$ and $s_j(t)$ is the spike state ($0$ or $1$) of neuron $j$.

Each Loihi neuron core supports arbitrary connection topologies as long as the capacities of the in-core memories for storing axons and synapses are not exceeded. The neuron cores are parallel and distributed with local on-chip SRAMs to store the network state and configurations. The neuron cores are fully asynchronous, performing synaptic accumulation only when there is an input event, which highly benefits from the spatio-temporal sparsity of event-based sensors and encoding. The algorithmic time step in the entire Loihi system is maintained by a distributed handshaking mechanism called \emph{barrier synchronization}. In addition, each Loihi chip has 3 synchronous embedded x86 cores also taking part in the barrier synchronization. The x86 cores run C code and are used to monitor and interact with the SNN running on the neuron cores, handling data IO between the on-chip asynchronous neuron cores and off-chip devices, and optionally synchronizing the algorithmic time steps duration (in physical time) to integrate the chip with a sensor.

We developed a solution to deploy the trained networks from our PyTorch implementation to Loihi (PyTorch2Loihi). First, we export the neurons' hyper-parameters and the trained synaptic weights from PyTorch in an HDF5 file by taking into account the Loihi hardware specifications and constraints as follows:

\begin{itemize}
    \item {\bf Current and voltage decays constants:} we calculate the Loihi decay constants $\delta^I$ and $\delta^U$ from the PyTorch time constants $\tau_{syn}$ and $\tau_{mem}$ respectively, with
    
    \begin{equation}
    \label{eq:loihi_decay}
        \delta=\mathrm{int} \left( 2^{-12} - 2^{-12} \cdot e^{\frac{-time\_bin\_size}{\tau}} \right)
    \end{equation}
    
    \item {\bf Synaptic weights and neuron threshold:} Loihi supports up to 8-bit fixed point weights. To minimize the effect of quantization, we quantize the weights from PyTorch training into 256 states and adjust the weight scaling factor and threshold scaling factor to have the same overall effect of an input spike. The weight scaling factor $w_{scale}$ and the weight quantization scheme is described by
    
    \begin{equation}
    \label{eq:loihi_scale}
        w_{scale}=\mathrm{int} \left({\textstyle\frac{256}{\max|w|}} \right)
    \end{equation}
    
    \begin{equation}
    \label{eq:loihi_weight}
        w_{Loihi}=\mathrm{quantize}\left(w, \,\mathrm{step}=2 \right) \cdot w_{scale}
    \end{equation}
    
    \begin{equation}
        \label{eq:loihi_threshold}
        \theta_{Loihi}=2^6 \cdot \theta \cdot w_{scale}
    \end{equation}
    
\end{itemize}

After the network is deployed, we run the inference with the event-based tactile data and first quantify the classification accuracy, then, measure the energy consumption and computation delay. The test inference was made by injecting the input events of all samples of the test set in a continuous flow, where samples are separated by a blank time of 100 algorithmic time steps where the neurons' currents and voltages decay to zero. The output spikes are gathered throughout the duration of the inference, and then the classification accuracy is calculated offline. 

Loihi system boards include voltage regulators and power telemetry which can be used to measure the total power consumption of the Loihi chip while a model is running. The power measurements can be combined with timing information recorded by the on-chip x86 cores during model operation to estimate energy consumption. NxSDK exposes a high-level user interface for measuring power, energy, and timing when a workload is running. We used the interface to benchmark the performance of our \gls{snn} models on Loihi.


\section{Results} \label{Results}

\subsection{Encoding analysis} \label{results:encoding_analysis}
To characterize the event-based datasets, we reconstructed the temporal sequences out of the event streams and compared the results with the original frame-based signal. Additionally, we performed the same analysis by taking into account the time binning step used to prepare the data for clock-driven computation as described in~\ref{methods:snn}. Results are summarized in Tab.~\ref{tab:encoding_analysis}, where the mean number of events, the compression ratio $\gamma$ with respect to the encoded data at $\vartheta=1$ and the reconstruction \gls{mse} values $\varepsilon$ are reported both, before and after the binning step for each threshold.

\subsubsection{Signal reconstruction before time binning} \label{encoding_analysis:reconstruction_encoding}
From the event stream, the signal was reconstructed, starting from zero, by increasing or decreasing, for every event, according to the polarity ON or OFF, by an amount equal to the threshold used in the encoding. Fig.~\ref{fig:encode_reconstruct}b shows the reconstruction values for one sample and taxel at different threshold values. The compression ratio $\gamma$ is defined as the number of events at $\vartheta=1$ divided by the number of events at each threshold value. The reconstruction error $\varepsilon$ is the \gls{mse} between the original sequence and the reconstructed frame-based sequence for each of the event-based datasets. \\
The analysis of the reconstructed frame-based signal revealed that with increasing thresholds the number of events dramatically decreases, increasing the reconstruction error. However, the compression ratio $\gamma$ increases with a higher rate than the reconstruction error $\varepsilon$, showing a sparsity gain of the event-based dataset at the cost of information content. \\
 
\begin{table}
    \caption{Characterisation of event-based encoding for each of the generated datasets at different threshold values.  Compression ratio $\gamma$ is defined as the number of events at perfect encoding ($\vartheta=1$) divided by the number of events at each higher threshold value. Values of the reconstruction error $\varepsilon$ are calculated per reconstructed frame by \gls{mse}. Mean events are calculated per sample. All $time\_bin\_size$, introduced to prepare the data for clock-driven computation, follow the results reported in~\ref{results:snn} from the two-step \gls{hpo} procedure.\\}
    \label{tab:encoding_analysis}
    \centering
    \begin{scriptsize}
        \begin{tabular}{>{\centering}m{0.10\linewidth}>{\centering}m{0.10\linewidth}>{\centering}m{0.10\linewidth}>{\centering}m{0.10\linewidth}>{\centering}m{0.10\linewidth}>{\centering}m{0.10\linewidth}>{\centering}m{0.10\linewidth}>{\centering\arraybackslash}m{0.10\linewidth}}
             & \multicolumn{3}{c}{\textbf{Before time binning}} & \multicolumn{4}{c}{\textbf{After time binning}} \\
             \cmidrule(lr){2-4}\cmidrule(lr){5-8}
	        \textbf{Threshold\\($\vartheta$)} & $\overline{\mathrm{\textbf{Events}}}$\\ & $\overline{\mathrm{\textbf{Comp. ratio}}}$\\($\gamma$) & \shortstack{$\overline{\mathrm{\textbf{MSE}}}$\\($\varepsilon$)} & \textbf{Bin size}\\(ms) & $\overline{\mathrm{\textbf{Events}}}$\\ & $\overline{\mathrm{\textbf{Comp. ratio}}}$\\($\gamma$) & \shortstack{$\overline{\mathrm{\textbf{MSE}}}$\\($\varepsilon$)} \\
	        \toprule
	        \textbf{1}  & 87.6 & 1    & 0 & \textbf{5} & 58.1	& 1.5	& 39.5\\
	        \midrule
	        \textbf{2}  & 38.0 & 2.3  &  0.4 & \textbf{3} & 35.5 & 2.5 & 12.5\\
	        \midrule
	        \textbf{5}  & 10.5 & 8.3  &  3.7 & \textbf{3} & 10.5 & 8.3 & 4.1\\
	        \midrule
	        \textbf{10} & 3.4 & 25.7  & 12.3 & \textbf{5} & 3.4 & 25.7 & 12.3\\
	        \bottomrule
        \end{tabular}
    \end{scriptsize}
\end{table}

\begin{figure}
    \centering
    \includegraphics[width=1.0\textwidth]{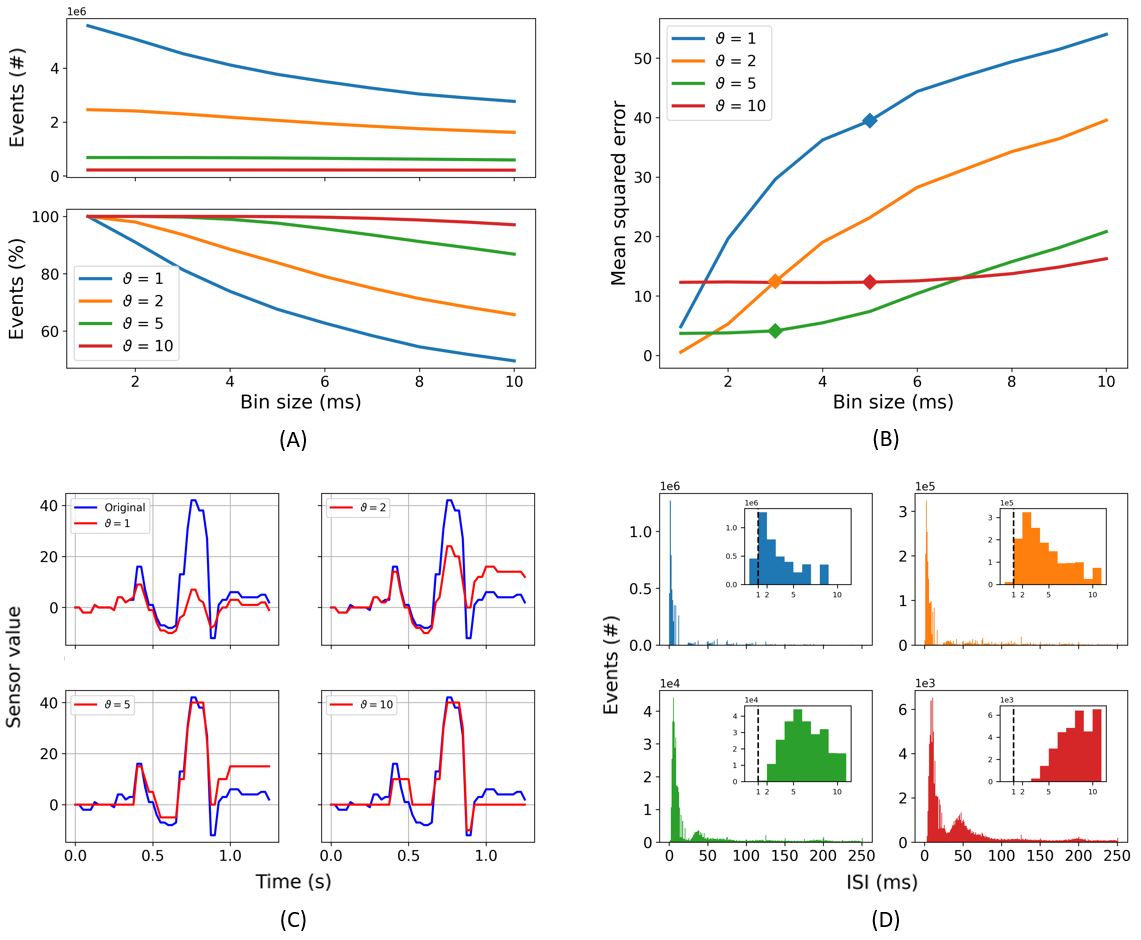}
    \caption{Spike encoding: (a) Total number of events counted in the whole dataset in dependence of the selected threshold and $time\_bin\_size$ is shown in the top panel, while the relative amount of events found in the dataset relative to $time\_bin\_size=1$ is reported in the bottom one. Increasing the encoding threshold reduces the number of events significantly, whereas encoding with lower thresholds is much more affected by increasing $time\_bin\_size$. The amount of events lost is 50.37~\% for $\vartheta=1$, 34.25~\% for $\vartheta=2$, 13.17~\% for $\vartheta=5$, and only 2.92~\% for $\vartheta=10$. (b) \gls{mse} values of signal reconstruction after time binning as a function of the $time\_bin\_size$ for each encoding threshold. The $time\_bin\_size$ resulting from the \gls{hpo} and grid search used to preprocess the event stream are highlighted by the markers. (c) Shows the reconstruction of the frame-based signal from the event stream for all given thresholds after time binning with a bin size of 5~ms. (d) Number of events as a function of the \gls{isi} with fixed $time\_bin\_size$ as reported in Tab.~\ref{tab:encoding_analysis}, with the same colour coding as in (a) and (b). The insets show the detail at \gls{isi} values equal to the $time\_bin\_size$ used in this work, highlighting the minimum temporal resolution of 1~ms with the vertical dashed line. Spikes below 1~ms in percentage of total number of spikes: 8.22~\% for $\vartheta=1$, 0.45~\% for $\vartheta=2$, 0~\% for $\vartheta=5$ and $\vartheta=10$.}
    \label{fig:encoding_analysis}
\end{figure}

\subsubsection{Signal reconstruction after time binning} \label{encoding_analysis:reconstruction_bin}
Running \glspl{snn} in PyTorch requires the introduction of time bins. To quantify the impact of the time binning on the different encoded datasets, we counted the total number of events given in each dataset after the time binning. The total number of events for a given encoding threshold is always higher than the total number of events for a given lower encoding threshold, regardless of the time binning, as shown in the top panel of Fig.~\ref{fig:encoding_analysis}a. The higher the encoding threshold, the lower the impact of the time binning. For the encoding threshold $\vartheta=1$, the total number of events counted in the dataset for a $time\_bin\_size$ equal to 1~ms is halved with increasing $time\_bin\_size$. For the encoding threshold $\vartheta=2$, we still have a loss of ~35~\%, whereas for higher encoding thresholds $\vartheta \geq 5$ the impact of the time binning decreases close to or below 10~\% and most of the events are perceived, as shown in the bottom panel of Fig.~\ref{fig:encoding_analysis}a.

The same reconstruction of the frame-based signal from the event stream as described in~\ref{encoding_analysis:reconstruction_encoding} was performed for every possible $time\_bin\_size$, included in our \gls{hpo} and grid search, of the event stream for each threshold value. The reconstruction error $\varepsilon$ depends, on the one hand, as discussed above, on the encoding threshold, and on the other hand on the introduced time binning. Results are shown in Fig.~\ref{fig:encoding_analysis}b with markers at the $time\_bin\_size$ selected after the \gls{hpo} and grid search to run the \glspl{snn} for each encoding threshold. The smallest reported reconstruction error $\varepsilon$ is 0.55 for threshold $\vartheta=2$ and 1~ms $time\_bin\_size$. We see a great increase in the reconstruction error $\varepsilon$ with increasing $time\_bin\_size$. That increase is even more drastic for threshold $\vartheta=1$ starting at a higher reconstruction error $\varepsilon$ of 4.82. The higher reconstruction error $\varepsilon$ can be explained by the loss of events when multiple of them fall into a single time bin, leading to an increase in the reconstruction error by introducing an accumulating offset, as shown in Fig.~\ref{fig:encoding_analysis}c. Additionally, the discriminative power of amplitudes is lost, leading to similar amplitude for small and high changes after the reconstruction. A further analysis unveiled, that 8.22~\% of the \glspl{isi} for threshold $\vartheta=1$ are below 1~ms which is the smallest $time\_bin\_size$ used, whereas for $\vartheta=2$ only 0.45~\% are below 1~ms, as summarized in Fig.~\ref{fig:encoding_analysis}d. For all higher thresholds ($\vartheta>2$) no \glspl{isi} are below 1~ms. The higher the threshold, the higher the majority of \glspl{isi} due to the increasing sparseness. The impact of time binning has decreasing impact, but the error introduced by the higher encoding thresholds is becoming more relevant. Overall, the higher thresholds are more resilient to the impact of $time\_bin\_size$, but less able to reflect the temporal dynamics.\\

\subsection{Standard classifiers for frame-based data} \label{results:standard_classifier}

\subsubsection{Linear classifier} \label{results:svm}

The recorded dataset encodes information in spatio-temporal patterns. Before this dataset is used for pattern recognition, it is important to know whether only spatial information or both spatial and temporal information in this dataset needs to be considered for character recognition. To investigate the significance of spatial and temporal information for the Braille letter classification task, we applied \gls{svm} with a linear kernel. To eliminate only the temporal information from the data by keeping the spatial information intact, we computed the mean of all frames with respect to time for each channel. In this way, we got one data point for each channel which is time independent and each channel represents the spatial location of each sensor, referred to as `time collapsed data'. Data keeping both, spatial and temporal information intact, is referred to as `raw data'. We applied a one-vs-rest multiclass classifier on the raw and the time-collapsed data for five cross-validated splits. We achieved (86.7~$\pm$~0.77)~\% and (72.3~$\pm$~1)~\% accuracy for raw and time-collapsed data respectively. To further investigate the temporal nature of the data we iteratively increased the number of frames taken into account from the first to the total number of frames in the frame-based data, in every iteration the data has been reduced to the first 12 principle components found by \gls{pca} to always consider the same dimensionality of the predictors in the classifier.

\begin{figure}
    \centering
    \includegraphics[width=1\textwidth]{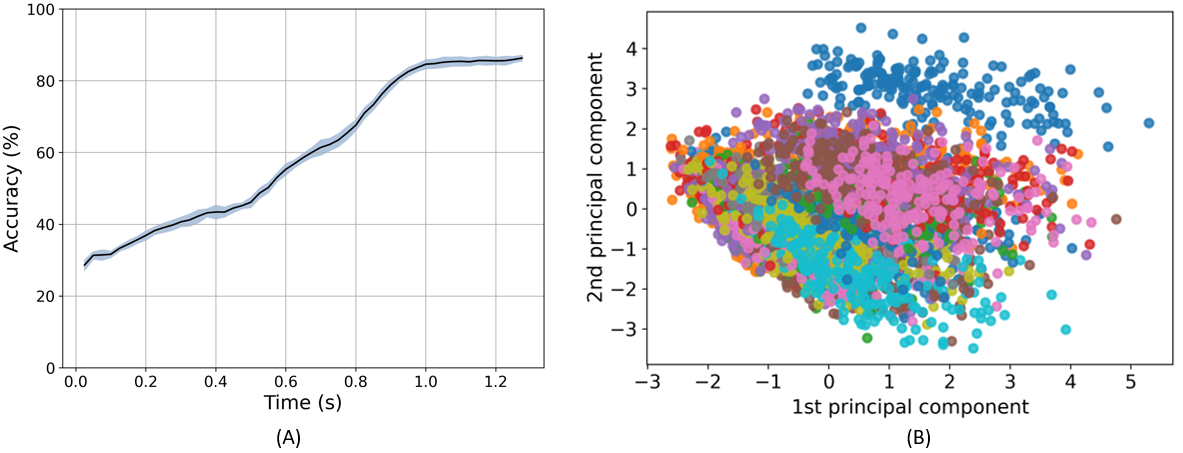}
    \caption{(a) Dependency of \gls{svm} performance in regard to the first 12 principle components extracted using \gls{pca} provided with an increasing number of frames. An increase in the number of frames results in an increase in the performance close to saturation in 1s at (86.7~$\pm$~0.77)~\%. (b) Dimensionality reduction (\gls{pca}) with 2 components applied to the spatio-temporal sequences of the frame-based dataset. Each of the colors in the visualization represents a category (letter) in the dataset.}
    \label{fig:pca_svm}
\end{figure}

In Fig.~\ref{fig:pca_svm}a the results of this procedure are shown, with a clear increase in the accuracy with three significant phases at 0.5~s and 0.8~s before it finally saturates after 1s at (86.7~$\pm$~0.77)~\%. We can identify the three phases in the sliding procedure, namely: the first contact with the dot pattern between 0.1~s and 0.5~s, the first contact with the second row of the pattern between 0.5~s and 0.8~s, and the end of the pattern after 1~s, by comparing Fig.~\ref{fig:fingertip_and_setup}c and Fig.~\ref{fig:pca_svm}a. 

Similarly, for the event-based dataset, we investigated the discrimination power of the spatial components by removing the temporal dimension of the dataset. In this case, we built the classifier with predictors as the summation of all the events for each taxel and channel (ON/OFF) resulting in 24 predictors per letter in the dataset. Fig.~\ref{fig:pca_svm}b shows the first two principal components after applying \gls{pca} to the spatial predictors displaying an overlapping in data categories. The result of the trained classifier drops from (58.94$\pm$1.15)~\% for the event-based data with all time bins as predictors to (47.7$\pm$1.48)~\% for the predictors that only account for the sum of spatial information. This analysis confirms the intuition that the temporal information of the signal is important for the character discrimination tasks.

\subsubsection{Time-series classifier and LSTM} \label{results:time_series_calssifier_and_lstm}

\begin{figure}
    \centering
    \includegraphics[width=0.9\textwidth]{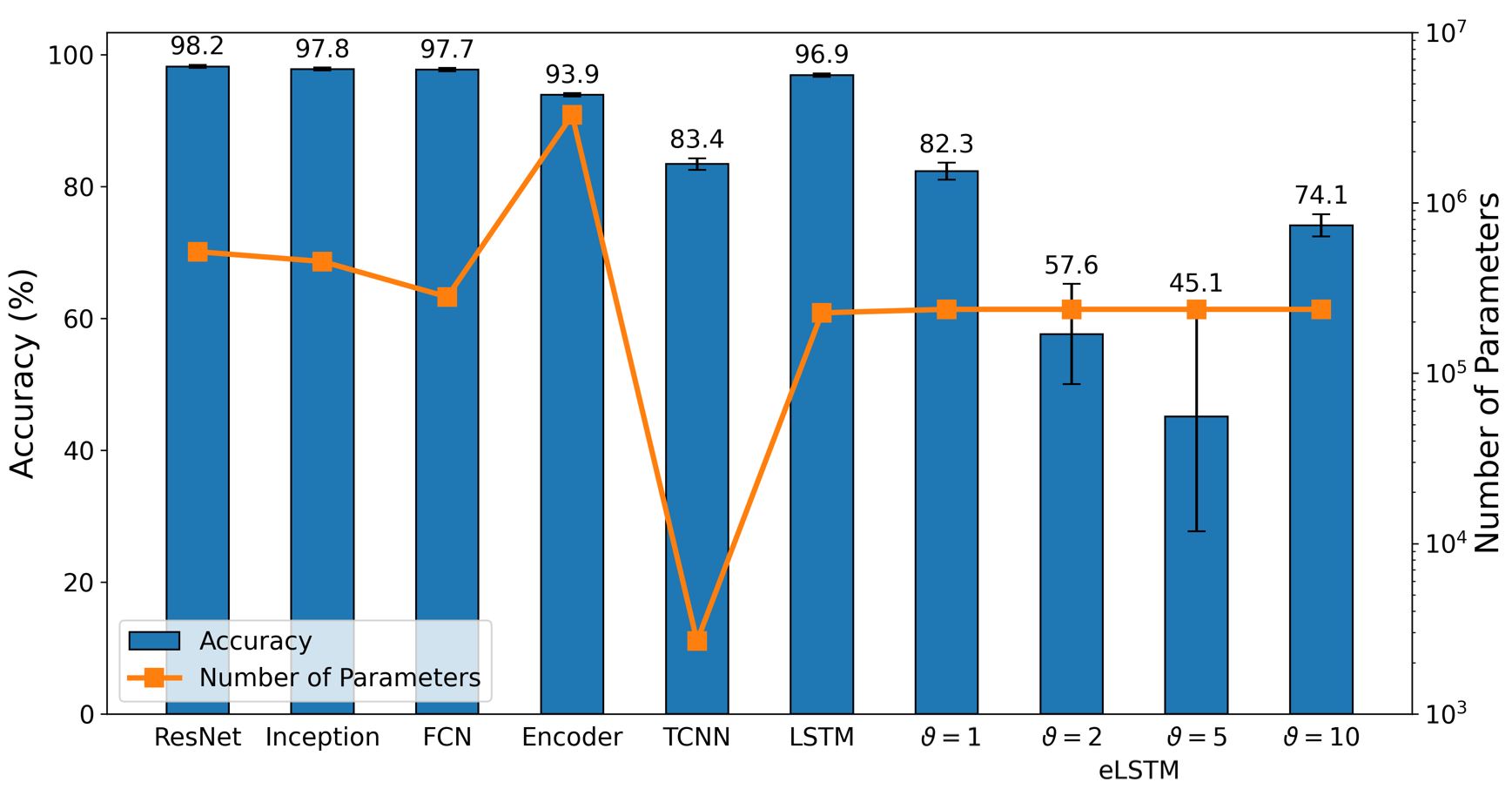}
    \caption{Test accuracy and number of trainable parameters of standard classifiers after training for 300 epochs and averaging over three runs. eLSTM refers to LSTM with event-based input.}
    \label{fig:accuracies_params_sota}
\end{figure}

We trained the time-series classifiers and the \gls{lstm} network with an 80/20 train-test split for 300 epochs and averaged the results over three runs per network. The resulting test accuracy as well as the network's respective number of trainable parameters are shown in Fig.~\ref{fig:accuracies_params_sota}.

\gls{resnet}, Inception and \gls{fcn} performed at comparable level. They all achieved $100~\%$ in training accuracy and reached comparable test accuracy of $(98.2~\pm~0.2)~\%$, $(97.8~\pm~0.2)~\%$ and $(97.7~\pm~0.3)~\%$, respectively. Similarly, their number of parameters turns out to be comparable too, with \gls{resnet} ($516k$) and Inception ($452k$) close to each other and \gls{fcn} ($280k$) slightly smaller but in the same order of magnitude.

The Encoder architecture reached a training accuracy of $(99.8~\pm~0.1)~\%$ but the drop in test accuracy was greater, resulting in $(93.9~\pm~0.2)~\%$. This deviation is an indication of overfitting and can possibly be explained by the fact that the network's number of parameters is one order of magnitude higher than the one of the former mentioned networks. This high amount of parameters can lead to a training outcome where the network memorizes the training data and conversely generalizes worse than it would with fewer parameters.

\gls{lstm} is the only recurrent architecture in our selection. It performs sequential processing from a streaming input without buffering the data. This reduces the delay and memory footprint which is important for embedded systems. Compared to the previous networks, it achieved again a training accuracy of 100~\% and the test accuracy did not drop as significantly as for the Encoder and resulted in $(96.9~\pm~0.3)~\%$, suggesting less overfitting.

\gls{tcnn} performed worst among all selected networks and only achieved $(88.5~\pm~0.8)~\%$ and $(83.4~\pm~1.3)~\%$ for training and test accuracy, respectively. However, it uses a greatly reduced number of parameters, containing only 2,673, which is two orders of magnitude lower than \gls{resnet}, Inception, \gls{fcn} and \gls{lstm}, and even three orders of magnitude lower than the Encoder network.

\subsection{LSTM with event-based input (eLSTM)} \label{results:lstm_events}

As a benchmark for the results provided by the optimized \glspl{rsnn}, \gls{lstm} was adopted for event-based data, using the same architecture as described in Section~\ref{methods:lstm}, the same train-test split as described in~\ref{results:time_series_calssifier_and_lstm} and the same data preprocessing as for the \gls{rsnn}, meaning sigma-delta encoding and time binning. Results are reported in Fig.~\ref{fig:accuracies_params_sota}. A clear impact of the encoding threshold on the e\gls{lstm} can be observed, with $\vartheta=1$ providing the best accuracy values. Compared to the optimized \glspl{rsnn}, while similar results are found when using $\vartheta=1$, significantly lower performances are achieved with $\vartheta=2$ and $\vartheta=5$, which also show a drastically increasing standard deviation. For $\vartheta=10$, results are again similar to those reported for the \glspl{rsnn}, but a higher standard deviation is observed in this case as well. A similar behaviour across the different threshold values can be observed in Fig.~\ref{fig:power_state_of_the_art} as well, where $\vartheta=2$ and $\vartheta=5$ show worse performances, in terms of inference time, compared to $\vartheta=1$ and $\vartheta=10$. One possible explanation is the same $time\_bin\_size$ for $\vartheta=1$ and $\vartheta=10$ with 5~ms, compared to 3~ms for the other encoding thresholds. This leads to $3/5$ the number of time steps to compute, which the \gls{lstm} seems to benefit from. \\

\subsection{Spiking neural networks} \label{results:snn}

During the two-step \gls{hpo} procedure, the main objective for the optimization was the classification accuracy, but we also monitored the \gls{ttc} and the power consumption, with the latter separately reported in Section~\ref{results:hardware}. In the perspective of an online implementation of the proposed \gls{rsnn}, an informative figure of merit to be accounted for is the minimum temporal length of the input needed for successful classification. To this aim, we defined \gls{ttc} as the portion of the signal needed for successful classification with respect to the full acquisition time of the Braille letter, which was fixed to 1.35~s, due to the fixed sliding speed. Additionally, in order to account for possible corner cases in the online implementation, like `no contact' situations, we included one extra class leading to 28 classes in total.

\begin{table}
    \caption{Optimized values of the hyperparameters, for each encoding threshold, following grid search.\\}
    \label{tab:hpo_results}
    \centering
    \begin{scriptsize}
        \begin{tabular}{>{\centering}m{0.18\linewidth}>{\centering}m{0.09\linewidth}>{\centering}m{0.09\linewidth}>{\centering}m{0.09\linewidth}>{\centering\arraybackslash}m{0.09\linewidth}}
	        \multirow{2.4}{*}{} & \multicolumn{4}{c}{\textbf{Threshold} ($\vartheta$)} \\
	        \cmidrule(lr){2-5}
	        & \textbf{1} & \textbf{2} & \textbf{5} & \textbf{10} \\
	        \toprule
	        \textbf{scale} & 5 & 15 & 10 & 10 \\
	        \midrule
	        \textbf{time\_bin\_size} \tiny{(ms)} & 5 & 3 & 3 & 5 \\
	        \midrule
	        \textbf{nb\_input\_copies} & 2 & 8 & 4 & 2 \\
	        \midrule
	        \textbf{tau\_mem} \tiny{(ms)} & 60 & 50 & 70 & 70 \\
	        \midrule
	        \textbf{tau\_ratio} & 10 & 10 & 10 & 10 \\
	        \midrule
	        \textbf{fwd\_weight\_scale} & 1 & 1 & 1.5 & 4 \\
	        \midrule
	        \textbf{weight\_scale\_factor} & 1e-2 & 2e-2 & 3.5e-2 & 1.5e-2 \\
	        \midrule
	        \textbf{reg\_spikes} & 4e-3 & 1.5e-3 & 1e-3 & 1.5e-3 \\
	        \midrule
	        \textbf{reg\_neurons} & 1e-6 & 0 & 0 & 0 \\
	        \bottomrule
        \end{tabular}
        
    \end{scriptsize}
\end{table}

The parameter space after the \gls{nni} optimization, as well as after the grid search, for each threshold, shows no significant trends. The complex interaction of different parameters leads to a variety of local optima resulting in comparable test accuracy. Looking only at the trials with the best classification accuracy for different encoding thresholds, reported in Tab.~\ref{tab:hpo_results}, gives the same picture. Only the forward weight scale, $fwd\_weight\_scale$, seems to be constantly increasing with increasing thresholds. The membrane potential time constant $\tau_{mem}$ has only slight variations, again with no clear trend, and the $tau\_ratio$, describing the relation of $\tau_{mem}$ and $\tau_{syn}$, is constant. Seeing similar membrane time constants indicates their dependencies on the spatial-temporal properties of the data despite the encoding threshold. A constant membrane-to-synapse time constant ratio in contrast shows to be an optimal relation between neuron and synapse dynamics for this task.

\begin{figure}
    \centering
    \includegraphics[width=1.0\textwidth]{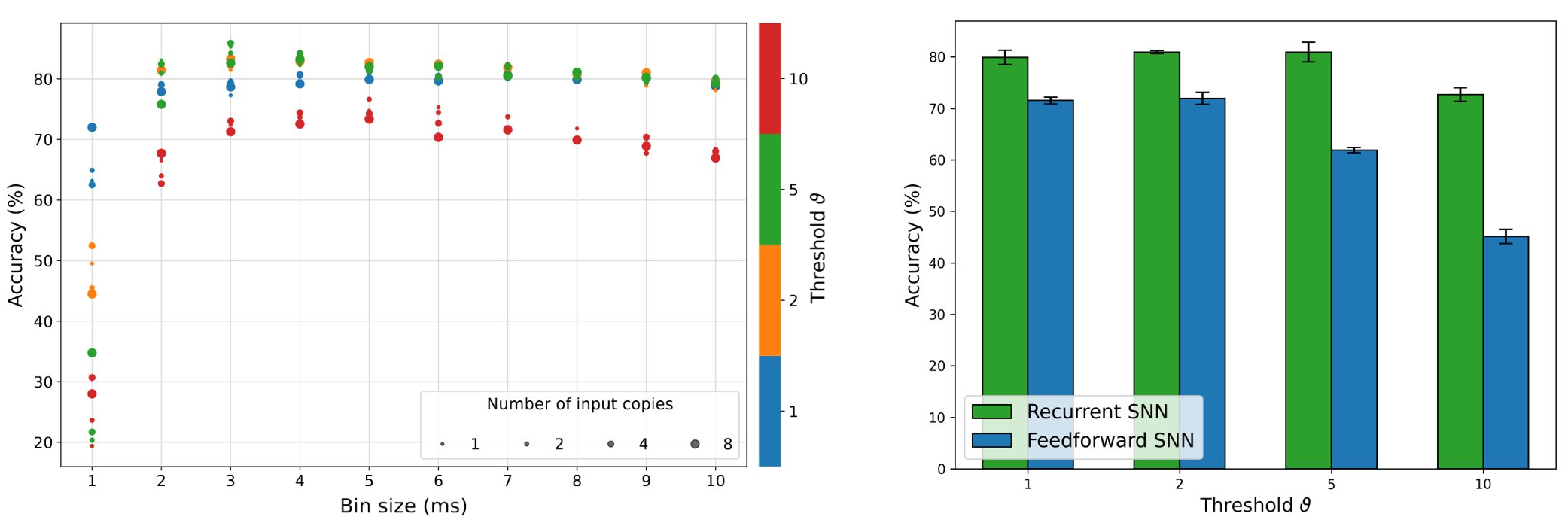}
    \caption{Summary of the accuracy performances of \gls{rsnn} and \gls{ffsnn} resulting from the grid search exploration in the two-step \gls{hpo} procedure. (a) Best test accuracy results achieved with the \gls{rsnn} for all the combinations of $time\_bin\_size$ and  $nb\_input\_copies$. (b) Mean and standard deviation of the accuracy results of both the \gls{ffsnn} and the \gls{rsnn} with the best parameters for each encoding threshold.}
    \label{fig:acc_plot}
\end{figure}

Fig.~\ref{fig:acc_plot} shows a summary of the \gls{rsnn} classification accuracy after grid search optimization, while in Supplementary Material~Fig.~S2 the hyperparameters exploration during the first step of the HPO procedure is reported. From all the explored combinations of $time\_bin\_size$ and $nb\_input\_copies$ for all the encoding thresholds employed, the best configuration in terms of accuracy turned out to be the one adopting an encoding threshold of $\vartheta=5$, a $time\_bin\_size$ of 3~ms and 4 input copies. Nevertheless, such \gls{rsnn} configuration did not provide the strongest reliability in terms of repeatability. As shown in Fig.~\ref{fig:acc_plot}b, the standard deviation of the provided results is larger than the one observed for other configurations. Fig.~\ref{fig:acc_plot}b shows that the best configuration found for an encoding threshold $\vartheta=2$ resulted in a mean accuracy as high as the one for an encoding threshold $\vartheta=5$ but with a significantly reduced standard deviation: (80.9~$\pm$~0.3)~\% test accuracy compared to (80.9~$\pm$~1.9)~\%. 

Looking at the accuracy performances from this twofold perspective, it is hence possible to identify as the best configuration the one employing the encoding threshold $\vartheta=2$ with a $time\_bin\_size$ of 3~ms and a number of copies equal to 8. The overall performance of the highest threshold ($\vartheta=10$) is the worst. The analysis of accuracy results provided an additional insight also, reported in both Fig.~\ref{fig:acc_plot}a and Fig.~\ref{fig:acc_plot}b, highlighting that the encoding threshold and $time\_bin\_size$ can have a significant impact. Particularly,  they induce a similar behavior of test accuracy: after an initial growth leading to a maximum, a further increase of one of them produces a deterioration of the classification performance. In contrast to the findings regarding the preservation of events concerning the $time\_bin\_size$ for different thresholds discussed in \ref{results:encoding_analysis}, the accuracy for higher encoding thresholds decreases most for higher $time\_bin\_size$. Comparing the network performance of the \gls{rsnn} and the \gls{ffsnn}, shown in Fig.~\ref{fig:acc_plot}b, we observe a constant decrease for the \gls{ffsnn} with increasing thresholds, but an almost constant performance for the \gls{rsnn} up to $\vartheta=10$, where it initially starts to drop off.

Targeting an online hardware implementation, next to classification performance, the energy efficiency needs to be considered too. From this point of view, the encoding threshold of 1 is the most promising candidate using a small number of input copies and a greater $time\_bin\_size$, leading to a significantly lower energy footprint at comparable performance. Regardless of the hyperparameters, the \gls{ttc} is equal throughout all conditions and the whole time series is needed for the best classification performance, similar to the findings using the linear \gls{svm}, reported in~\ref{results:svm}. To validate the use of the extra class, we also compared these results with an implementation based on 27 classes only. Such analysis revealed that similar results are achieved in both cases, thus showing that the additional 28th class does not have a detrimental impact on the classification performances.

\subsection{Hardware implementation} \label{results:hardware}

\subsubsection{NVIDIA Jetson embedded GPU} \label{results:hardware_jetson}
\begin{figure}
    \centering
    \includegraphics[width=0.9\textwidth]{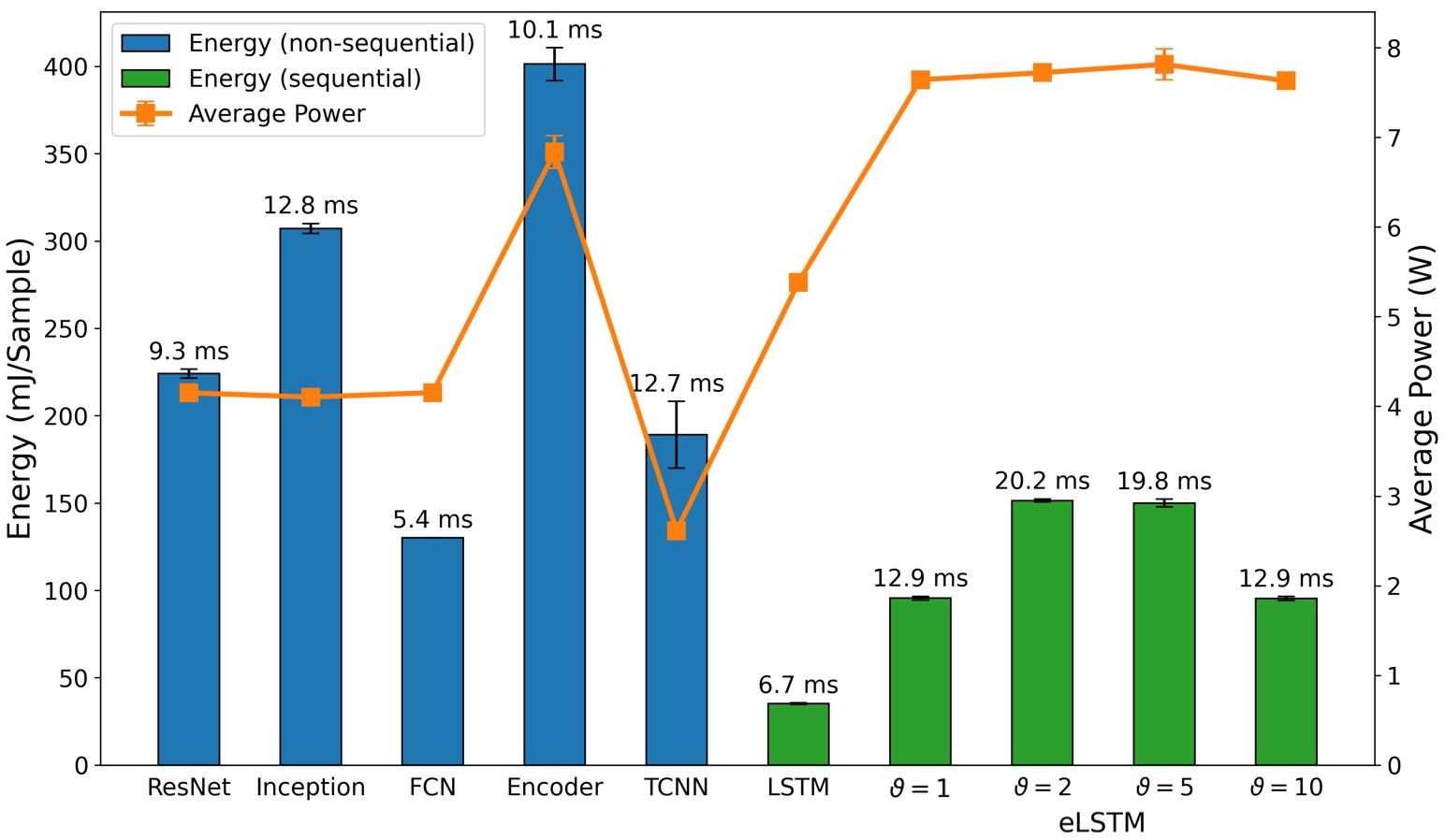}
    \caption{Comparison of inference metrics from the standard classifiers for frame-based data in terms of energy consumption and average power usage as measured on an NVIDIA Jetson Xavier NX. eLSTM refers to LSTM with event-based input. The label on top of each bar shows the inference time per sample on the corresponding network.}
    \label{fig:power_state_of_the_art}
\end{figure}

The energy consumption, average power usage, and inference time of the standard classifiers running on the NVIDIA Jetson\footnote{Energy and timing measurements for the Jetson were obtained on an NVIDIA Jetson Xavier NX Developer Kit running the NVIDIA JetPack 4.6.1 SDK, containing Ubuntu 18.04.6 LTS, L4T 32.7.1, TensorRT 8.2.1, and CUDA 10.2} are shown in Fig.~\ref{fig:power_state_of_the_art}. We consider energy consumption the main metric as it includes both power usage and inference time per sample. Average power usage gives insight into what power budget would be required to achieve certain inference times. While energy and power consumption are important metrics for battery-powered applications, inference time is crucial in applications with real-time constraints.

When comparing these results with the parameter counts in Fig.~\ref{fig:accuracies_params_sota}, some similarities can be found. \gls{resnet}, Inception, and \gls{fcn}, which are comparable in terms of accuracy and parameter count, had nearly the same average power usage during inference. This means that their energy consumption is directly proportional to their inference time. In the case of Inception and its parameter count, we expected the energy consumption to lie between \gls{resnet} and \gls{fcn}, but it exceeded both. While parameter count is not directly related to computational complexity, another explanation for this observation could be that Inception uses operations that are less optimized or not accelerated by the GPU. The similarities continue with the Encoder, where average power usage jumped and the parameter count increased by one order of magnitude, compared to the former three networks. Judging only by the parameter count, we expected energy consumption to be even higher. The difference, especially compared to Inception, is not as great as expected, though. Considering the higher average power usage, it had better GPU utilization and therefore could benefit from an overall higher acceleration. Inference time also supports this assumption, as it was lower on the Encoder than on Inception.

Amongst all standard classifiers, the results for the \gls{lstm} with frame-based input are the most notable. Despite being a sequential network, which we expected to be slower compared to the standard classifiers due to its iterative and recurrent nature, it achieved the fastest inference time and lowest energy consumption. But this came at the cost of the second highest power consumption. Similar to the Encoder, this suggests that the network can benefit from a high GPU utilization or general acceleration of internal operations. 

For the \gls{lstm} with the event-based input, further referred to as eLSTM, we found that the energy per sample and the inference time reflect the number of time steps processed with a ratio of 5 to 3 which is given by the time binning applied for each encoding threshold. Similar $time\_bin\_size$ for $\vartheta=1$ and $\vartheta=10$ with 5~ms resulting in 270 time steps and for $\vartheta=2$ and $\vartheta=5$ with 3~ms resulting in 450 time steps have been used. Interestingly this ratio does not hold when compared to the frame-based signal with 54 timesteps. The reason for that might be the nature of the data with float numbers for the frame-based data and integer numbers for the event stream. The power consumption for all eLSTMs is nearly the same.

Lastly, the \gls{tcnn} performed poorly when putting it alongside the other networks. It had, on average, a similar energy consumption but a high inference time, despite having by far the least amount of parameters of the shown networks. In general, it seems that this specific architecture is not very well suited for solving the problem at hand.

\begin{figure}
    \centering
    \includegraphics[width=0.9\textwidth]{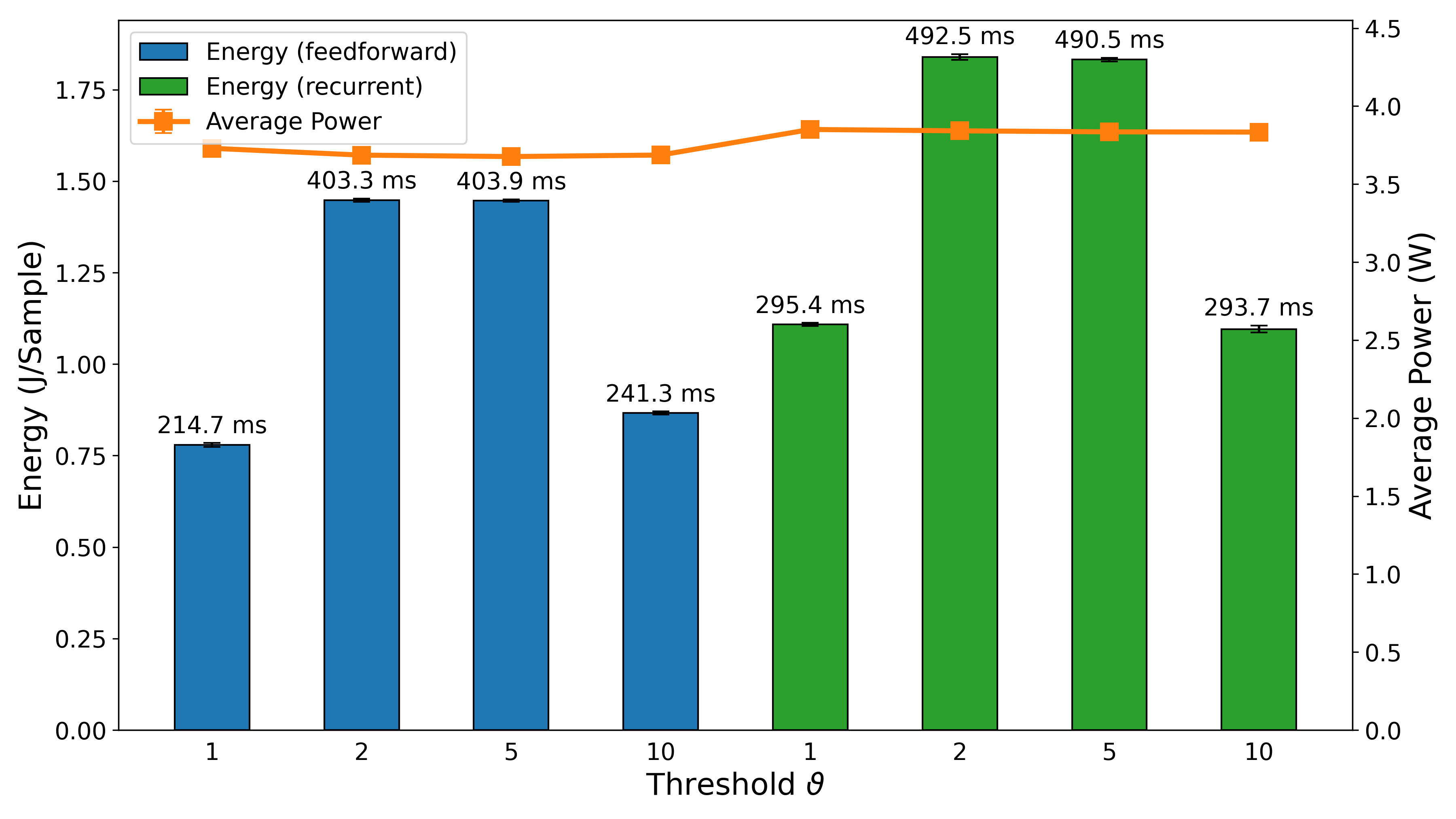}
    \caption{Comparison of inference metrics for all spiking neural networks in terms of energy consumption and average power usage as measured on an NVIDIA Jetson Xavier NX. The label on top of each bar shows the inference time per sample on the corresponding network.}
    \label{fig:power_spiking}
\end{figure}

For the \glspl{snn} and \glspl{rsnn}, we expected results on the NVIDIA Jetson to be solely proportional to the parameters $time\_bin\_size$ and $nb\_input\_copies$, and whether the feedforward or recurrent architecture was used. These three factors mainly define the number of operations to be calculated during inference. Conversely, we assumed that the threshold does not affect performance as the implementation on general-purpose computers does not take advantage of the temporal sparsity in the data. Our measurements generally support these assumptions and are shown in Fig.~\ref{fig:power_spiking}. The average power usage of the feedforward and recurrent architectures across their thresholds were very close, deviating less than 0.7~\% for the former, and less than 0.2~\% for the latter from their respective means. This implies a constant utilization of computational resources as well as the energy consumption being directly proportional to the inference time for each architecture. Thresholds $\vartheta=1$ and $\vartheta=10$ as well as thresholds $\vartheta=2$ and $\vartheta=5$ consumed roughly the same amount of energy per inference. Tab.~\ref{tab:hpo_results} shows that both threshold pairs presumably depend on $time\_bin\_size$ and $nb\_input\_copies$. The only exception is $nb\_input\_copies$ for threshold $\vartheta=2$ and $\vartheta=5$, which is 8 and 4, respectively. This brings us to the conclusion that $nb\_input\_copies$ does not have a meaningful impact on energy consumption and inference time in real-life scenarios, whereas $time\_bin\_size$ and the type of architecture are the main factors for the computational load. A comparison between the \glspl{snn} and the eLSTMs shows an increased inference time of 20$\times$, even though both compute the same number of time steps, indicating the lack of computational optimization for the \glspl{snn} computation. The energy per sample is 10$\times$ higher and the average power consumption 1.5$\times$.

In general, when comparing the absolute numbers of Fig.~\ref{fig:power_spiking} with the standard classifiers in Fig.~\ref{fig:power_state_of_the_art}, our implementations of the \gls{ffsnn} and \gls{rsnn} have a clear disadvantage at energy consumption and inference time when being run on a GPU accelerated device. The most efficient \gls{snn} consumed $\sim$88~\% more energy than the least efficient standard classifier, and the fastest spiking network took 16.8$\times$ longer for one inference than the slowest standard classifier. These numbers clearly show the need for dedicated neuromorphic hardware to implement event-based algorithms.

\subsubsection{Intel Loihi neuromorphic chip} \label{results:hardware_loihi}

\begin{figure}
    \centering
    \includegraphics[width=0.7\textwidth]{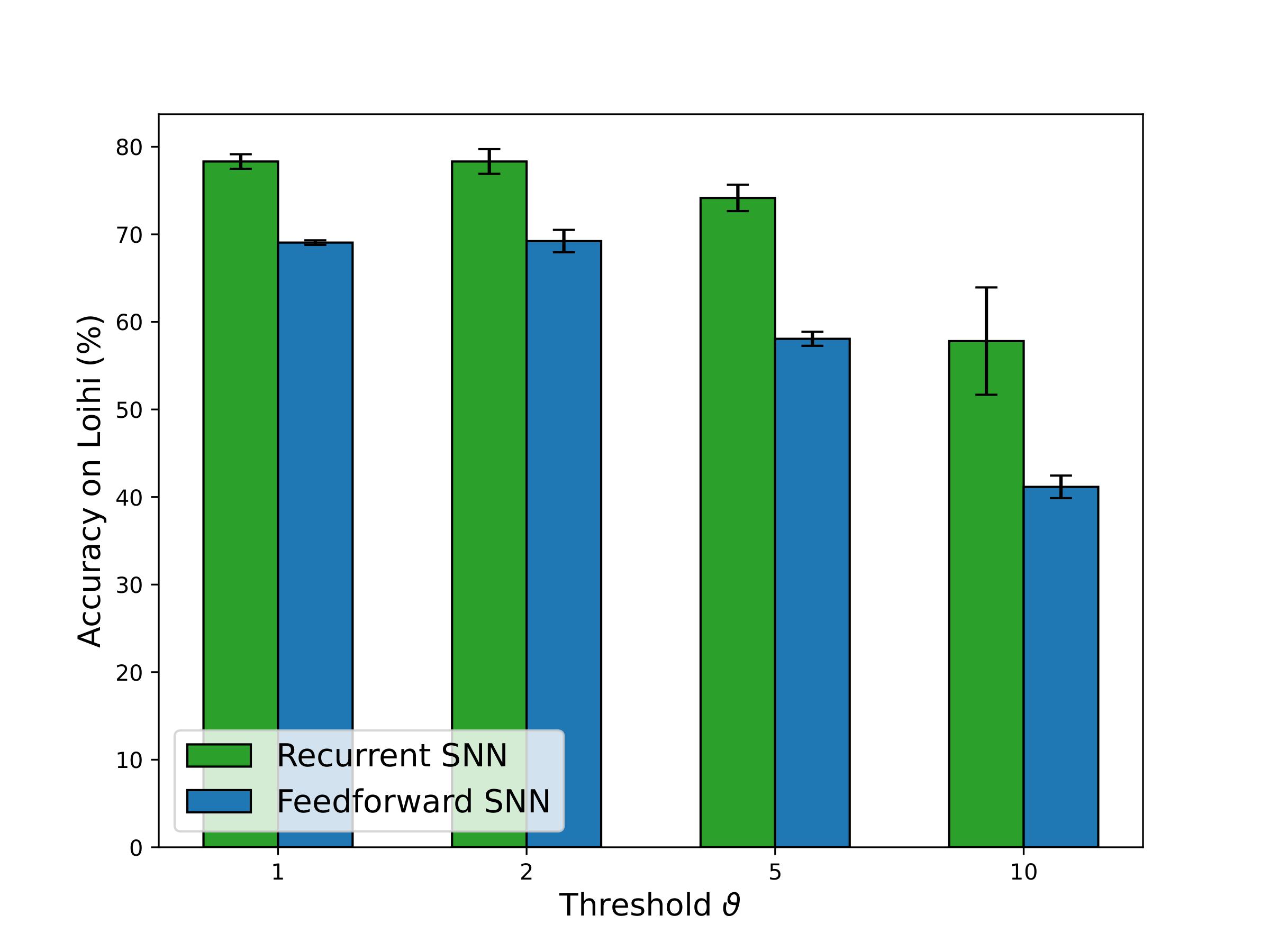}
    \caption{Comparison of the accuracy results for \gls{ffsnn} and \gls{rsnn} on Loihi with the best parameters found by the two-stage \gls{hpo} for each encoding threshold.}
    \label{fig:loihi_accuracy}
\end{figure}

The overall trend of accuracy in Loihi for the \glspl{snn}, shown in Fig.~\ref{fig:loihi_accuracy} follows the trend of accuracy on the PyTorch simulations and Jetson inference shown in Fig.~\ref{fig:acc_plot}b. Nevertheless, there is a loss in accuracy of a few percent (e.g. -1.58~\% for the \gls{rsnn} with encoding threshold $\vartheta=1$, which is due to the PyTorch training procedure without accounting for the Loihi hardware constraints, in particular, the 8-bit fixed point weights implementation. The loss varies depending on the PyTorch weights distribution.

\begin{figure}
    \centering
    \includegraphics[width=0.9\textwidth]{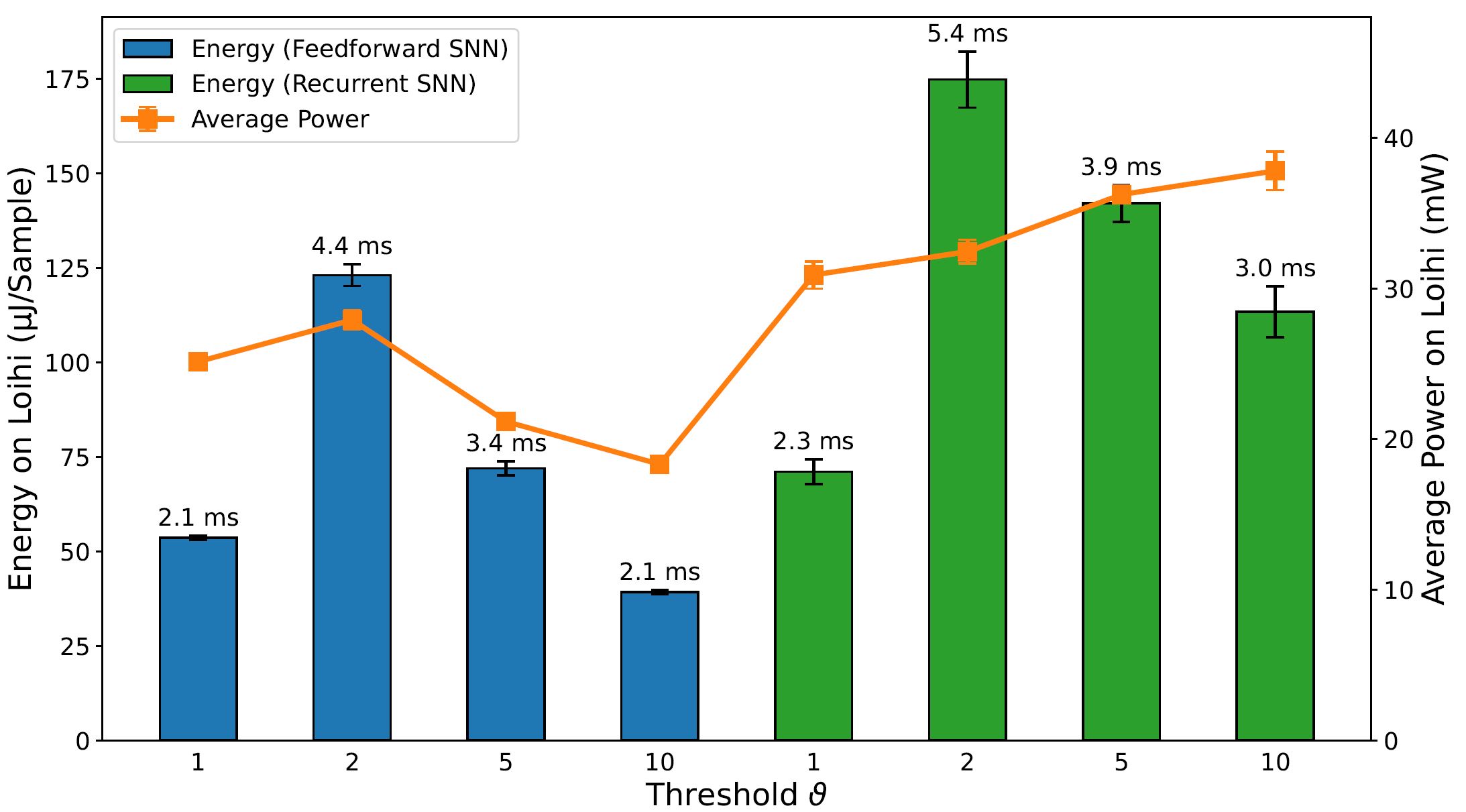}
    \caption{Comparison of inference metrics for all trained spiking neural networks in terms of energy consumption and average power usage as measured on Loihi. The label on top of each bar shows the inference time per sample on the corresponding network.}
    \label{fig:loihi_energy}
\end{figure}

Then, we compared the hardware efficiency of the recurrent and \gls{ffsnn} in terms of delay (i.e. execution time), power and energy consumption. Before discussing the results, it is important to specify the neural cores mapping we have used which does not affect the accuracy but does affect the hardware efficiency. Loihi offers flexibility in how to map the network neurons into the neural cores, constrained by the number of cores in a chip as well as the number of input axons, synapses, neurons, and output axons in a neural core. The goal is then to find a good trade-off between parallelism (i.e. using more neural cores with fewer neurons per core) and time-multiplexing (i.e. using fewer neural cores with more neurons per core), to balance the neural core's power, the mesh routing power and algorithmic time step duration to get an optimal configuration for the application requirements. We found, that for the specific typologies, the more cores we use the more power and energy we consume without a significant impact on the delay. We, therefore, used the minimum number of cores considering all the hardware constraints which are 8 cores for all the trained networks. 

 The delay, power and energy consumption of the different deployed \glspl{snn} on Loihi\footnote{Energy and timing measurements were obtained on Nahuku 32 board ncl-ext-ghrd-01 with an Intel(R) Xeon(R) CPU E5-2650 0 @ 2.00~GHz and 4GB RAM running Ubuntu 20.04.4 LTS and NxSDK v1.0.0} are shown in Fig.~\ref{fig:loihi_energy}. For the sake of simplicity, we refer to energy consumption as the main metric, as it includes both power and delay. As expected, \glspl{ffsnn} consumes less energy than \glspl{rsnn}, for the same thresholds. This is due to the overhead of memory and computation from the recurrent synaptic connections. \glspl{ffsnn} and \glspl{rsnn}  follow a similar trend for the different thresholds: networks with thresholds $\vartheta=2$ and $\vartheta=5$ consume more energy because they have smaller bin sizes and therefore more time bins (i.e. algorithmic time steps) per sample (450) compared to networks with thresholds $\vartheta=1$ and $\vartheta=10$ (270), and they have more input copies as shown in Tab.~\ref{tab:hpo_results}. Networks with threshold $\vartheta=2$ consume more than networks with threshold $\vartheta=5$, mainly because they have more input copies (8~vs.~5). Nevertheless, the \gls{ffsnn} with threshold $\vartheta=1$ consumes more than the \gls{ffsnn} with threshold $\vartheta=10$, while the \gls{rsnn} with threshold $\vartheta=1$ consumes less than the \gls{rsnn} with threshold $\vartheta=10$. Even though in both cases the networks with threshold $\vartheta=1$ have more events in the input and fewer events in the hidden layer than the networks with threshold $\vartheta=10$, shown in the Supplementary Material~Tab.~S2, the impact of the hidden layer events is different, because every event in the \gls{ffsnn} hidden layers gets transmitted to the 28 output neurons while every event in the \glspl{rsnn} hidden layers gets transmitted to both the 28 output neurons and the 450 hidden neurons. Therefore, the gain obtained in the input layer for the \gls{rsnn} with threshold $\vartheta=10$ is lost with the recurrent topology which increases the number of synaptic operations. Finally, while the Jetson GPU is mostly sensitive to the number of bins as shown in Fig.~\ref{fig:power_spiking}, Loihi is also sensitive to the number of input copies and the spatio-temporal sparsity of the spikes and synaptic operations in the network.

\begin{table}
    \caption{Results summary from \gls{rsnn} on Loihi and \gls{rsnn}, eLSTM and \gls{lstm} on Jetson for accuracy, total power, energy per sample, delay and energy-delay product. The number of trainable parameter (i.e. synaptic weights) are similar between the \gls{rsnn} (236,700), the \gls{lstm} (225,975) and the eLSTM (236,919). Event-based inputs are encoded with threshold $\vartheta=1$. Comparisons with respect to RSNN on Loihi are evaluated as differences for the accuracy and as ratios for all the other quantities.\\}
    \label{tab:loihi_comparison}
    \centering
    \begin{scriptsize}
        \begin{tabular}{>{\raggedright}m{0.15\linewidth}>{\centering}m{0.09\linewidth}>{\centering}m{0.09\linewidth}>{\centering}m{0.09\linewidth}>{\centering}m{0.09\linewidth}>{\centering}m{0.09\linewidth}>{\centering}m{0.09\linewidth}>{\centering\arraybackslash}m{0.09\linewidth}}
             & \multicolumn{4}{c}{\textbf{Results summary}} & \multicolumn{3}{c}{\textbf{Comparison with RSNN on Loihi}} \\
             \cmidrule(lr){2-5}\cmidrule(lr){6-8}
	        \textbf{Network} & \textbf{RSNN} & \textbf{RSNN} & \textbf{eLSTM} & \textbf{LSTM} & \textbf{RSNN} & \textbf{eLSTM} & \textbf{LSTM} \\
	        \textbf{Hardware} & Loihi & Jetson & Jetson & Jetson & Jetson & Jetson & Jetson \\
	        \textbf{Input} & Events & Events & Events & Frames & Events & Events & Frames \\
	        \toprule
	        \textbf{Accuracy} (\%) & 78.32 & 79.90 & 82.31 & 96.92 & +1.58 & +3.99 & +18.60 \\
	        \midrule
	        \textbf{Total power} (mW) & 31 & 3,851 & 7,642 & 5,385 & 124$\times$ & 247$\times$ & 174$\times$ \\
	        \midrule
	        \textbf{Total energy\\per sample} ($\mu$J) & 71 & 1,108,695 & 96,000 & 35,212 & 15,615$\times$ & 1,352$\times$ & 496$\times$ \\
	        \midrule
	        \textbf{Delay per sample} (ms) & 2.3 & 295.3 & 12.9 & 6.7 & 172$\times$ & 5.6$\times$ & 2.9$\times$ \\
	        \midrule
	        \textbf{Energy-delay product} ($\mu$J$\cdot$s) & 0.16 & 327,398 & 1,238 & 236 & 2,046,237$\times$ & 7,738$\times$ & 1,475$\times$ \\
	        \bottomrule
        \end{tabular}
    \end{scriptsize}
\end{table}

After the comparison of the different deployed \glspl{snn} on Loihi, taking into account the accuracy, power, energy, and delay, we conclude that the \gls{rsnn} with encoding threshold $\vartheta=1$ is the best option. For simplicity, we will refer to it as the \gls{rsnn} for the rest of this section. 

Tab.~\ref{tab:loihi_comparison} compares \gls{rsnn} in Loihi, the \gls{rsnn} on Jetson, the \gls{lstm} and the eLSTM on Jetson. The \gls{rsnn} on Loihi loses 1.58~\% in accuracy compared to the \gls{rsnn} on Jetson, mainly due to the quantization that was done after training. It further loses 17~\% compared to the \gls{lstm} on Jetson, but only 3~\% compared to the eLSTM. A specific \gls{lstm} architecture with the same number of parameters as in the \gls{rsnn} has been used and trained for 300 epochs. However, the \gls{rsnn} on Loihi show several orders of magnitude gains in hardware efficiency. First, compared to the \gls{rsnn} on Jetson, it is 124$\times$ more power-efficient and 172$\times$ faster, which makes it four orders of magnitude (15,615$\times$) more energy-efficient. It clearly shows that \glspl{snn} are particularly inefficient when implemented on standard GPU hardware. It should nevertheless be highlighted that the delay or execution time of the \gls{rsnn} on Jetson still satisfies the real-time constraint imposed by the sensor which has a sampling frequency of 40~Hz (i.e. a maximum algorithmic time step duration of 25~ms). Even though the average execution time of the \gls{rsnn} on Jetson is relatively long (295.38~ms), it is still lower than the total duration of each sample (1,350~ms). It is to note that this delay can increase in the real-world setting when adding off-chip communication with the robot. Second, compared to the \gls{lstm} on Jetson, the \gls{rsnn} on Loihi is more than 170$\times$ more power-efficient and exhibits a 2.9$\times$ longer average execution time, which makes it three orders of magnitude (1,475$\times$) more energy-efficient.

Finally, compared to the standard eLSTM classifier on the Jetson GPU using the same event data, the neuromorphic approach with the Loihi chip and \glspl{rsnn} is about 4~\% less accurate but two orders of magnitude (247$\times$) more power-efficient, reducing the total power from 7.642~W to about 31~mW. Furthermore, as a consequence of the reduced execution time, from 12.9~ms down to 2.3~ms, the neuromorphic approach introduces a gain in energy efficiency of 1,352$\times$ and a gain in energy-delay product of 7,738$\times$. This goes in line with recent results of \glspl{snn} on Loihi compared to standard algorithms and hardware, where the best performing workloads on Loihi make use of highly recurrent networks \cite{Davies_etal21}. Furthermore, we can expect an even higher gain in energy efficiency when using the \gls{rsnn} on Loihi in the real-world environment, because it exploits the spatio-temporal sparsity of the event-driven encoding. Therefore, if the robot is not moving its finger, no event is transmitted to the Loihi chip, drastically reducing the dynamic power that represents about 20~mW out of the total 31~mW. Instead, the Jetson GPU would always process the redundant frames coming from the sensor. Our study shows how event-driven encoding, neuromorphic hardware, and \glspl{snn} are put together to improve the overall efficiency of tactile pattern recognition, emphasizing the importance of the neuromorphic approach for embedded applications with a continuous stream of data.


\section{Conclusions} \label{Conclusions}
The initial analysis of the frame-based data using a linear classifier demonstrates that the information in our dataset is encoded in both the spatial and the temporal domains. The results show a decrease in the accuracy of linear classifiers when no time dimension is accounted for in both frame- and event-based data, thus motivating the use of architectures that are capable to learn spatio-temporal patterns from the data. Although linear classifiers provide a high accuracy when all time bins are taken into account as predictors, it is not the desired approach to learning spatio-temporal patterns since 
they cannot be applied online where data needs to be gathered in real-time instead of being already available. This is in contrast with the sequential learning approaches and spike-based algorithms explored in this manuscript.

The data encoding analysis presents the trade-off between information content from the original frame-based data and the sparsity of the event stream. The original frame-based data is inherently redundant since the information content decreases slower than the compression ratio of events revealed by comparing the datasets at different threshold levels. Next to the impact of the encoding threshold, the analysis has shown that time binning needs to be considered as an important influencing factor. Due to the need to create vectors with sparse event representation to run on GPUs, the smallest $time\_bin\_size$ creates the lowest boundary in which the temporal dynamics of the event stream can be correctly represented. If \glspl{isi} fall below the $time\_bin\_size$ information content and temporal dynamics suffer and information is lost. Based on the implementation results, the network may not reflect the sparsity in the input event stream at every network layer. An optimized network for task performance showed an increase in the number of events and quantity of energy consumption of the total architecture although the dataset presented to the input layer has a compression ratio bigger than 1 in terms of events. It is true, that the number of time bins has a significant impact on the power consumption, but comparing threshold $\vartheta = 1$ with $\vartheta = 10$ as well as $\vartheta = 2$ with $\vartheta = 5$ for the \gls{rsnn}, which have the same $time\_bin\_size$ of 5~ms and 3~ms respectively, the higher threshold has in both cases a higher power consumption as highlighted in our hardware implementation on Loihi shown in Fig.~\ref{fig:loihi_energy}, which can only be explained by a higher number of spikes transmitted in the network as reported in the Supplementary Material~Tab.~S2. Therefore, encoding schemes that minimize the number of events in the input may result in higher energy consumption in the system as a whole.

We further found that the spiking neuron current and voltage time constants that were optimized independently for each encoding threshold are similar because they are correlated to the inherent temporal dynamics of the input event stream rather than the encoding threshold or binning time window. The \gls{hpo} did not settle at a global optimum for any of the encoding thresholds, which underlines the highly complex interaction of the included parameters, leading to many locally optimal solutions. Looking at the results of the follow-up grid search reveals, that all $time\_bin\_size$ greater than 2~ms lead to a similar trend showing a slight decrease in classification accuracies for increasing $time\_bin\_size$. Relating the moderate decrease of accuracy with the increase in energy and power saving makes the selection of higher $time\_bin\_size$ for further robotic implementation preferable.

Our findings regarding the implementation of the \glspl{snn} on the NVIDIA Jetson show that it is capable of fulfilling the constraints of real-time performance, which is defined here as the inference time of any network is lower than the recording time of one sample. The increase of the inference time between non-spiking and spiking architectures is substantial, though. It ranges from $\sim16\times$ (Inception vs. \gls{ffsnn} with threshold $\vartheta=1$) to $\sim91\times$ (\gls{fcn} vs. \gls{rsnn} with threshold $\vartheta=2$). As a consequence, the energy consumption also rose by a significant margin which ranges from $\sim2\times$ (Encoder vs. \gls{ffsnn} with threshold $\vartheta=1$) to $\sim40\times$ (\gls{lstm} vs. \gls{rsnn} with threshold $\vartheta=2$). These numbers show the clear drawback of our implementation on conventional hardware. For a meaningful deployment, it would either need a more optimized implementation that is better accelerated by GPUs, or dedicated hardware that can take advantage of the characteristics of the spiking domain, like temporal sparsity.

We have been able to demonstrate the possibility of efficiently performing time series classification by exclusively using event-based encoding and asynchronous event-based computation on neuromorphic hardware. Our deployed \gls{rsnn} can discriminate between 27 classes of Braille letters with $78.32$~\% accuracy, using only 450 recurrently connected hidden units and consuming a total amount of $31$~mW on the Intel Loihi neuromorphic chip. This is yet not sufficient to report a competitive classification performance compared to standard classifiers or other results achieved with \glspl{snn} on different tasks. The encoding analysis revealed that too much information is lost using sigma-delta encoding and that hinders the network from further performance improvements. Nevertheless, comparing these findings to a \gls{lstm} on the NVIDIA Jetson embedded GPU yields a gain in power-efficiency of $250\times$ for the \gls{rsnn} with threshold $\vartheta=2$ on Loihi. On the one hand, these results show the challenges of spike-based computing compared to standard algorithms in terms of accuracy; on the other hand, they highlight the opportunities of the neuromorphic approach with event-based communication and asynchronous processing in terms of power/energy efficiency and delay, especially for mobile robots or highly energy-constrained fields of applications. In addition to the neuromorphic computation while performing the task, event-based encoding is important in this context. Event-based systems can be considered at rest while no significant change occurs and with that, the power consumption during that time is extremely low. Nevertheless, the system can react to changes immediately, due to its asynchronous nature. 

The \gls{lstm} with the frame-based data-stream outperforms the \gls{rsnn} in accuracy by $17$~\%, whereas the eLSTM achieves comparable performance to the the \gls{rsnn}, highlighting the need for better event-based encoding techniques. For example, graded spikes supported by Loihi 2 \cite{Orchard_etal21} could reduce the information loss by adding the magnitude of change to the event (when it exceeds a threshold), keeping both event-based communication and precise information. Furthermore, models mimicking the biological skin with its wide range of neuronal responses like the slow and fast adaptive receptors \cite{abraira_13} can be beneficial in terms of information extracted from the stimuli using multiple parallel pathways. Regarding the \gls{snn}, several mechanisms can be explored to improve the pattern recognition performance of the \glspl{rsnn}. First, by only applying a single recurrent hidden layer, we limited the trainable parameters in our network and with that, its learning capabilities. Future investigations might use multiple hidden layers to increase the spatio-temporal encoding power. Second, at the neuron level, we used homogeneous time constants for all neurons. Learning the time constants along with the synaptic weights to adapt the different neurons' temporal dynamics has been shown to be beneficial for the classification performance~\cite{Perez-Nieves_etal21}. Third, recent works suggest the need for more powerful recurrent units for spiking neurons to bridge the gap with \gls{lstm} and other formal (i.e. non-spiking) recurrent neuron models \cite{He_etal20,ParedesValles_etal21}. Finally, on another note, in contrast to this offline benchmarking methodology, an online classification could be performed by implementing a winner-take-all network \cite{Chen17} as the output layer or by using the recent attentive \gls{rsnn} with a decision-making circuit \cite{Yin_etal22}, enabling the network's prediction readout at each time step.

Overall, we presented a tactile spatio-temporal dataset suitable for benchmarking event-based encoding schemes and neuromorphic algorithms. We compared an event-based encoding and spike-based learning algorithm with standard machine learning approaches and implemented a classification system into different hardware platforms showing the advantages of the end-to-end neuromorphic approach for tactile pattern recognition. Nevertheless, our findings are not limited to tactile data, because spatio-temporal information is present in all sensory modalities when processed in a streaming fashion at the edge, and our approach can be applied to these types of workloads. Building full neuromorphic systems that inherently sense, process and communicate their output in an event-based manner provide great potential in terms of energy efficiency and scalability for embedded and \textit{embodied} sensory-motor systems that interact in real-time with the real-world environment. Despite the energy benefits of neuromorphic hardware, neuromorphic algorithmic space is still being explored, therefore, not as mature as standard \gls{ann} methods. We envision that this problem and the performance metrics we used will serve as a benchmark to drive progress in the field.


\section*{Conflict of Interest Statement} \label{Conflict of Interest Statement}
The authors declare that the research was conducted in the absence of any commercial or financial relationships that could be construed as a potential conflict of interest.


\section*{Author Contribution} \label{Author Contribution}
SMC had the initial idea and recorded the dataset. APZ and SMC performed the frame-to-event conversion. VF, EF, and DK performed the NNI implementation. VF, DK, and APZ run the optimization. VF and SMC implemented the performance evaluation. APZ, DK, SMC, VF, and DI investigated standard classifiers and carried out the encoding analysis. APZ, SMC, and SR investigated linear SVM and obtained PCA plots. DK acquired GPU measurements. LK deployed the \glspl{snn} on Loihi and acquired the measurements. LK, GU, FZ, and CB supervised the work. All authors contributed to the manuscript writing and approved the submitted version.


\section*{Acknowledgement} \label{Acknowledgement}
The authors would like to acknowledge the 2021 Telluride neuromorphic workshop and all its participants for the fruitful discussions, and Intel Corporation for access to the Loihi neuromorphic platform. The authors would like to thank Sumit Bam Shrestha from Intel Neuromorphic Computing Lab for his support of the Loihi implementation and feedback on the manuscript. The research activity herein was partially carried out using the HPC infrastructure as well as the NVIDIA Jetson platform of the Silicon Austria Labs (SAL), the HPC infrastructure of the IIT, and the Peregrine HPC cluster of the University of Groningen.


\section*{Funding} \label{Funding}
This work has been supported by European Union's Horizon 2020 MSCA Programme under Grant Agreement No 813713 NeuTouch, the “University SAL Labs” initiative of Silicon Austria Labs (SAL) and its Austrian partner universities for applied fundamental research for electronic based systems, the CogniGron research center and the Ubbo Emmius Funds of the University of Groningen.


\section*{Data Availability Statement} \label{Data Availability Statement}
The code for encoding, training, inference, and exporting the HDF5 file from PyTorch is open source and available in a \href{https://github.com/event-driven-robotics/tactile_Braille_reading}{GitHub repository}. The code for importing the HDF5 into Loihi is shared within the Intel Neuromorphic Research Community (INRC). The \href{https://zenodo.org/record/7050094}{dataset} is open access and uploaded in Zenodo under the DOI 10.5281/zenodo.7050094~\cite{muller_cleve_simon_f_2022_6556273}.


\bibliographystyle{IEEEtran}
\bibliography{bibliography}

\begin{thebibliography}{10}
\providecommand{\url}[1]{#1}
\csname url@samestyle\endcsname
\providecommand{\newblock}{\relax}
\providecommand{\bibinfo}[2]{#2}
\providecommand{\BIBentrySTDinterwordspacing}{\spaceskip=0pt\relax}
\providecommand{\BIBentryALTinterwordstretchfactor}{4}
\providecommand{\BIBentryALTinterwordspacing}{\spaceskip=\fontdimen2\font plus
\BIBentryALTinterwordstretchfactor\fontdimen3\font minus
  \fontdimen4\font\relax}
\providecommand{\BIBforeignlanguage}[2]{{%
\expandafter\ifx\csname l@#1\endcsname\relax
\typeout{** WARNING: IEEEtran.bst: No hyphenation pattern has been}%
\typeout{** loaded for the language `#1'. Using the pattern for}%
\typeout{** the default language instead.}%
\else
\language=\csname l@#1\endcsname
\fi
#2}}
\providecommand{\BIBdecl}{\relax}
\BIBdecl

\bibitem{romo2001touch}
R.~Romo and E.~Salinas, ``Touch and go: decision-making mechanisms in
  somatosensation,'' \emph{Annual review of neuroscience}, vol.~24, no.~1, pp.
  107--137, 2001.

\bibitem{prescott2011activetouch}
\BIBentryALTinterwordspacing
T.~J. Prescott, M.~E. Diamond, and A.~M. Wing, ``Active touch sensing,''
  \emph{Philosophical Transactions of the Royal Society B: Biological
  Sciences}, vol. 366, no. 1581, pp. 2989--2995, 2011. [Online]. Available:
  \url{https://royalsocietypublishing.org/doi/abs/10.1098/rstb.2011.0167}
\BIBentrySTDinterwordspacing

\bibitem{Bach-y-Rita04}
P.~{Bach-y-Rita}, ``Tactile sensory substitution studies,'' \emph{Annals of the
  New York Academy of Sciences}, vol. 1013, 2004.

\bibitem{martiniello2020association}
N.~Martiniello and W.~Wittich, ``The association between tactile, motor and
  cognitive capacities and braille reading performance: a scoping review of
  primary evidence to advance research on braille and aging,'' \emph{Disability
  and Rehabilitation}, pp. 1--15, 2020.

\bibitem{bola2016Braille}
L.~Bola, K.~Siuda-Krzywicka, M.~Papli{\'n}ska, E.~Sumera, P.~Ha{\'n}czur, and
  M.~Szwed, ``Braille in the sighted: Teaching tactile reading to sighted
  adults,'' \emph{PloS one}, vol.~11, no.~5, p. e0155394, 2016.

\bibitem{brysbaert2019many}
M.~Brysbaert, ``How many words do we read per minute? a review and
  meta-analysis of reading rate,'' \emph{Journal of Memory and Language}, vol.
  109, p. 104047, 2019.

\bibitem{kawabe2019experimental}
H.~Kawabe, S.~Seto, H.~Nambo, and Y.~Shimomura, ``Experimental study on
  scanning of degraded braille books for recognition of dots by machine
  learning,'' in \emph{International Conference on Management Science and
  Engineering Management}.\hskip 1em plus 0.5em minus 0.4em\relax Springer,
  2019, pp. 322--334.

\bibitem{li2010optical}
J.~Li and X.~Yan, ``Optical braille character recognition with support-vector
  machine classifier,'' in \emph{2010 International Conference on Computer
  Application and System Modeling (ICCASM 2010)}, vol.~12.\hskip 1em plus 0.5em
  minus 0.4em\relax IEEE, 2010, pp. V12--219.

\bibitem{hsu2020Braille}
B.-M. Hsu, ``Braille recognition for reducing asymmetric communication between
  the blind and non-blind,'' \emph{Symmetry}, vol.~12, no.~7, p. 1069, 2020.

\bibitem{shokat2020deep}
S.~Shokat, R.~Riaz, S.~S. Rizvi, A.~M. Abbasi, A.~A. Abbasi, and S.~J. Kwon,
  ``Deep learning scheme for character prediction with position-free touch
  screen-based braille input method,'' \emph{Human-centric Computing and
  Information Sciences}, vol.~10, no.~1, pp. 1--24, 2020.

\bibitem{li2014deep}
T.~Li, X.~Zeng, and S.~Xu, ``A deep learning method for braille recognition,''
  in \emph{2014 International Conference on Computational Intelligence and
  Communication Networks}.\hskip 1em plus 0.5em minus 0.4em\relax IEEE, 2014,
  pp. 1092--1095.

\bibitem{bartolozzi2016robots}
C.~Bartolozzi, L.~Natale, F.~Nori, and G.~Metta, ``Robots with a sense of
  touch,'' \emph{Nature materials}, vol.~15, no.~9, pp. 921--925, 2016.

\bibitem{bartolozzi2017event}
C.~Bartolozzi, P.~M. Ros, F.~Diotalevi, N.~Jamali, L.~Natale, M.~Crepaldi, and
  D.~Demarchi, ``Event-driven encoding of off-the-shelf tactile sensors for
  compression and latency optimisation for robotic skin,'' in \emph{2017
  IEEE/RSJ International Conference on Intelligent Robots and Systems
  (IROS)}.\hskip 1em plus 0.5em minus 0.4em\relax IEEE, 2017, pp. 166--173.

\bibitem{lichtsteiner2008128}
P.~Lichtsteiner, C.~Posch, and T.~Delbruck, ``A 128$\times$128 120 db 15 $\mu$s
  latency asynchronous temporal contrast vision sensor,'' \emph{IEEE journal of
  solid-state circuits}, vol.~43, no.~2, pp. 566--576, 2008.

\bibitem{conradt2009embedded}
J.~Conradt, R.~Berner, M.~Cook, and T.~Delbruck, ``An embedded aer dynamic
  vision sensor for low-latency pole balancing,'' in \emph{2009 IEEE 12th
  International Conference on Computer Vision Workshops, ICCV Workshops}.\hskip
  1em plus 0.5em minus 0.4em\relax IEEE, 2009, pp. 780--785.

\bibitem{chan2007aer}
V.~Chan, S.-C. Liu, and A.~van Schaik, ``Aer ear: A matched silicon cochlea
  pair with address event representation interface,'' \emph{IEEE Transactions
  on Circuits and Systems I: Regular Papers}, vol.~54, no.~1, pp. 48--59, 2007.

\bibitem{leonard1993tidigits}
R.~G. Leonard and G.~Doddington, ``Tidigits speech corpus,'' \emph{Texas
  Instruments, Inc}, 1993.

\bibitem{friedl2016human}
K.~E. Friedl, A.~R. Voelker, A.~Peer, and C.~Eliasmith, ``Human-inspired
  neurorobotic system for classifying surface textures by touch,'' \emph{IEEE
  Robotics and Automation Letters}, vol.~1, no.~1, pp. 516--523, 2016.

\bibitem{rongala2015neuromorphic}
U.~B. Rongala, A.~Mazzoni, and C.~M. Oddo, ``Neuromorphic artificial touch for
  categorization of naturalistic textures,'' \emph{IEEE transactions on neural
  networks and learning systems}, vol.~28, no.~4, pp. 819--829, 2015.

\bibitem{see2020stmnist}
\BIBentryALTinterwordspacing
H.~See, B.~Lim, S.~Li, H.~Yao, W.~Cheng, H.~Soh, and B.~C.~K. Tee, ``{ST-MNIST}
  - the spiking tactile {MNIST} neuromorphic dataset,'' \emph{CoRR}, vol.
  abs/2005.04319, 2020. [Online]. Available:
  \url{https://arxiv.org/abs/2005.04319}
\BIBentrySTDinterwordspacing

\bibitem{bologna2013closed}
L.~Bologna, J.~Pinoteau, J.~Passot, J.~Garrido, J.~Vogel, E.~R. Vidal, and
  A.~Arleo, ``A closed-loop neurobotic system for fine touch sensing,''
  \emph{Journal of neural engineering}, vol.~10, no.~4, p. 046019, 2013.

\bibitem{pinoteau2012closed}
J.~Pinoteau, L.~L. Bologna, J.~A. Garrido, and A.~Arleo, ``A closed-loop
  neurorobotic system for investigating braille-reading finger kinematics,'' in
  \emph{International Conference on Human Haptic Sensing and Touch Enabled
  Computer Applications}.\hskip 1em plus 0.5em minus 0.4em\relax Springer,
  2012, pp. 407--418.

\bibitem{jamali2015icubfingertip}
N.~Jamali, M.~Maggiali, F.~Giovanniniand, G.~Metta, and L.~Natale, ``A new
  design of a fingertip for the icub hand,'' \emph{IEEE/RSJ International
  Conference on Intelligent Robots and Systems (IROS)}, 2015.

\bibitem{rosa_2011}
J.~M. de~la Rosa, ``Sigma-delta modulators: Tutorial overview, design guide,
  and state-of-the-art survey,'' \emph{IEEE Transactions on Circuits and
  Systems I: Regular Papers}, vol.~58, no.~1, pp. 1--21, 2011.

\bibitem{wang2017ijcnn}
Z.~Wang, W.~Yan, and T.~Oates, ``Time series classification from scratch with
  deep neural networks: A strong baseline,'' in \emph{2017 International Joint
  Conference on Neural Networks (IJCNN)}, 2017, pp. 1578--1585.

\bibitem{geng2018cscnn}
\BIBentryALTinterwordspacing
Y.~Geng and X.~Luo, ``Cost-sensitive convolution based neural networks for
  imbalanced time-series classification,'' \emph{CoRR}, vol. abs/1801.04396,
  2018. [Online]. Available: \url{http://arxiv.org/abs/1801.04396}
\BIBentrySTDinterwordspacing

\bibitem{santiago2018encoder}
\BIBentryALTinterwordspacing
J.~Serr{\`{a}}, S.~Pascual, and A.~Karatzoglou, ``Towards a universal neural
  network encoder for time series,'' \emph{CoRR}, vol. abs/1805.03908, 2018.
  [Online]. Available: \url{http://arxiv.org/abs/1805.03908}
\BIBentrySTDinterwordspacing

\bibitem{zhao2017cnntimeseries}
B.~Zhao, H.~Lu, S.~Chen, J.~Liu, and D.~Wu, ``Convolutional neural networks for
  time series classification,'' \emph{Journal of Systems Engineering and
  Electronics}, vol.~28, no.~1, pp. 162--169, 2017.

\bibitem{hassen2020inceptiontime}
\BIBentryALTinterwordspacing
H.~I. Fawaz, B.~Lucas, G.~Forestier, C.~Pelletier, D.~F. Schmidt, J.~Weber,
  G.~I. Webb, L.~Idoumghar, P.-A. Muller, and F.~Petitjean, ``Inceptiontime:
  Finding alexnet for time series classification,'' \emph{Data Mining and
  Knowledge Discovery}, vol.~34, no.~1, pp. 162--169, 11 2020. [Online].
  Available: \url{https://doi.org/10.1007/s10618-020-00710-y}
\BIBentrySTDinterwordspacing

\bibitem{ismail2019deeplearning}
\BIBentryALTinterwordspacing
H.~I. Fawaz, G.~Forestier, J.~Weber, L.~Idoumghar, and P.-A. Muller, ``Deep
  learning for time series classification: a review,'' \emph{Data Mining and
  Knowledge Discovery}, vol.~33, no.~4, p. 917–963, Mar 2019. [Online].
  Available: \url{http://dx.doi.org/10.1007/s10618-019-00619-1}
\BIBentrySTDinterwordspacing

\bibitem{hochreiter1997lstm}
\BIBentryALTinterwordspacing
S.~Hochreiter and J.~Schmidhuber, ``{Long Short-Term Memory},'' \emph{Neural
  Computation}, vol.~9, no.~8, pp. 1735--1780, 11 1997. [Online]. Available:
  \url{https://doi.org/10.1162/neco.1997.9.8.1735}
\BIBentrySTDinterwordspacing

\bibitem{cramer2020heidelberg}
B.~Cramer, Y.~Stradmann, J.~Schemmel, and F.~Zenke, ``The heidelberg spiking
  data sets for the systematic evaluation of spiking neural networks,''
  \emph{IEEE Transactions on Neural Networks and Learning Systems}, pp. 1--14,
  2020.

\bibitem{zenke2021remarkable}
\BIBentryALTinterwordspacing
F.~Zenke and T.~P. Vogels, ``The remarkable robustness of surrogate gradient
  learning for instilling complex function in spiking neural networks,''
  \emph{Neural Computation}, vol.~33, no.~4, pp. 899--925, Mar 2021. [Online].
  Available: \url{https://doi.org/10.1162/neco\_a\_01367}
\BIBentrySTDinterwordspacing

\bibitem{neftci2019surrogategradient}
E.~O. Neftci, H.~Mostafa, and F.~Zenke, ``Surrogate gradient learning in
  spiking neural networks: Bringing the power of gradient-based optimization to
  spiking neural networks,'' \emph{IEEE Signal Processing Magazine}, vol.~36,
  no.~6, 2019.

\bibitem{adam2017autodiff}
A.~Paszke, S.~Gross, S.~Chintala, G.~Chanan, E.~Yang, Z.~DeVito, Z.~Lin,
  A.~Desmaison, L.~Antiga, and A.~Lerer, ``Automatic differentiation in
  pytorch,'' in \emph{NIPS Autodiff Workshop}.\hskip 1em plus 0.5em minus
  0.4em\relax Conference on Neural Information Processing System, 2017, pp.
  1--4.

\bibitem{fra2022human}
\BIBentryALTinterwordspacing
V.~Fra, E.~Forno, R.~Pignari, T.~C. Stewart, E.~Macii, and G.~Urgese, ``Human
  activity recognition: suitability of a neuromorphic approach for on-edge
  {AIoT} applications,'' \emph{Neuromorphic Computing and Engineering}, vol.~2,
  no.~1, p. 014006, Feb. 2022. [Online]. Available:
  \url{https://doi.org/10.1088/2634-4386/ac4c38}
\BIBentrySTDinterwordspacing

\bibitem{forno2019parallel}
E.~Forno, A.~Acquaviva, Y.~Kobayashi, E.~Macii, and G.~Urgese, ``{A Parallel
  Hardware Architecture For Quantum Annealing Algorithm Acceleration},'' in
  \emph{2018 IFIP/IEEE International Conference on Very Large Scale Integration
  (VLSI-SoC)}, vol. 2018-Octob.\hskip 1em plus 0.5em minus 0.4em\relax IEEE,
  2018, pp. 31--36.

\bibitem{davies2018loihi}
M.~Davies, N.~Srinivasa, T.-H. Lin, G.~Chinya, Y.~Cao, S.~H. Choday, G.~Dimou,
  P.~Joshi, N.~Imam, S.~Jain \emph{et~al.}, ``Loihi: A neuromorphic manycore
  processor with on-chip learning,'' \emph{Ieee Micro}, vol.~38, no.~1, pp.
  82--99, 2018.

\bibitem{Davies_etal21}
M.~Davies, A.~Wild, G.~Orchard, Y.~Sandamirskaya, G.~A.~F. Guerra, P.~Joshi,
  P.~Plank, and S.~R. Risbud, ``Advancing neuromorphic computing with loihi: A
  survey of results and outlook,'' \emph{Proceedings of the IEEE}, vol. 109,
  no.~5, pp. 911--934, 2021.

\bibitem{Orchard_etal21}
\BIBentryALTinterwordspacing
G.~Orchard, E.~P. Frady, D.~B.~D. Rubin, S.~Sanborn, S.~B. Shrestha, F.~T.
  Sommer, and M.~Davies, ``Efficient neuromorphic signal processing with loihi
  2,'' 2021. [Online]. Available: \url{https://arxiv.org/abs/2111.03746}
\BIBentrySTDinterwordspacing

\bibitem{abraira_13}
V.~E. Abraira and D.~D. Ginty, ``The sensory neurons of touch,'' \emph{Neuron},
  vol.~79, 2013.

\bibitem{Perez-Nieves_etal21}
N.~Perez-Nieves, V.~Leung, P.~Dragotti, and D.~Goodman, ``Neural heterogeneity
  promotes robust learning,'' \emph{Nature Communications}, vol.~12, p. 5791,
  10 2021.

\bibitem{He_etal20}
W.~He, Y.~Wu, L.~Deng, G.~Li, H.~Wang, Y.~Tian, W.~Ding, W.~Wang, and Y.~Xie,
  ``Comparing snns and rnns on neuromorphic vision datasets: Similarities and
  differences,'' \emph{Neural networks : the official journal of the
  International Neural Network Society}, vol. 132, pp. 108--120, 2020.

\bibitem{ParedesValles_etal21}
F.~Paredes-Vall{\'e}s, J.~J. Hagenaars, and G.~C. de~Croon, ``Self-supervised
  learning of event-based optical flow with spiking neural networks,'' in
  \emph{NeurIPS}, 2021.

\bibitem{Chen17}
\BIBentryALTinterwordspacing
Y.~Chen, ``Mechanisms of winner-take-all and group selection in neuronal
  spiking networks,'' \emph{Frontiers in Computational Neuroscience}, vol.~11,
  2017. [Online]. Available:
  \url{https://www.frontiersin.org/article/10.3389/fncom.2017.00020}
\BIBentrySTDinterwordspacing

\bibitem{Yin_etal22}
\BIBentryALTinterwordspacing
B.~Yin, Q.~Guo, F.~Corradi, and S.~Bohte, ``Attentive decision-making and
  dynamic resetting of continual running srnns for end-to-end streaming keyword
  spotting,'' in \emph{Proceedings of the International Conference on
  Neuromorphic Systems 2022}, ser. ICONS '22.\hskip 1em plus 0.5em minus
  0.4em\relax New York, NY, USA: Association for Computing Machinery, 2022.
  [Online]. Available: \url{https://doi.org/10.1145/3546790.3546795}
\BIBentrySTDinterwordspacing

\bibitem{muller_cleve_simon_f_2022_6556273}
S.~F. Müller-Cleve, ``Tactile braille letters dataset,'' Distributed by Zenodo
  {https://doi.org/10.5281/zenodo.6556273}, May 2022.

\end{thebibliography}

\end{document}